\documentclass[final,1p,times,number]{elsarticle}

\usepackage{amssymb}
\usepackage{mdframed}  
\usepackage[hyphens]{url}       
 
 \usepackage{multirow}
\usepackage{wrapfig}
\usepackage{fontawesome5}
\usepackage{tabularx}
\usepackage{hyperref}  
\usepackage{caption}         
\usepackage{subcaption}      
\usepackage{booktabs}        
\usepackage{array}           
\usepackage{amsmath}
\usepackage{booktabs}
\usepackage{adjustbox}
\usepackage{enumitem}       
\usepackage{amsmath}        
\usepackage{amssymb}        
\usepackage{textcomp}       

 \usepackage{example}
\newcommand{\issue}[1]{\textcolor{red}{#1}} 

\usepackage{color} 
\definecolor{mygreen}{rgb}{0.0, 0.6, 0.0}
\definecolor{myred}{rgb}{0.8, 0.0, 0.0}

\usepackage{graphicx}
\usepackage{array}
\usepackage[table,xcdraw]{xcolor}
\usepackage{pifont}
\usepackage{fontawesome5}

\journal{Elsevier}

\begin{document}

\begin{frontmatter}



\title{\underline{VLDBench} Evaluating Multimodal Disinformation with Regulatory Alignment}



\author[1]{Shaina Raza\corref{cor1}}
\ead{shaina.raza@vectorinstitute.ai}
\author[2]{Ashmal Vayani}
\author[3]{Aditya Jain}
\author[1]{Aravind Narayanan}
\author[1]{Vahid Reza Khazaie}
\author[4]{Syed Raza Bashir}
\author[5]{Elham Dolatabadi}
\author[5]{Gias Uddin}
\author[6]{Christos Emmanouilidis}
\author[2]{Rizwan Qureshi}
\author[2]{Mubarak Shah}

\cortext[cor1]{Corresponding author}

\affiliation[1]{organization={Vector Institute for Artificial Intelligence},
            addressline={MaRS Centre}, 
            city={Toronto},
            postcode={ON M5G 1L7}, 
            state={Ontario},
            country={Canada}}

\affiliation[2]{organization={Center for Research in Computer Vision, University of Central Florida},
            addressline={4000 Central Florida Blvd},
            city={Orlando},
            postcode={FL 32816}, 
            state={Florida},
            country={USA}}

\affiliation[3]{organization={Independent Researcher},
            city={USA}}

\affiliation[4]{organization={Toronto Metropolitan University},
            addressline={350 Victoria Street},
            city={Toronto},
            postcode={ON M5B 2K3},
            state={Ontario},
            country={Canada}}

\affiliation[5]{organization={York University},
            addressline={4700 Keele Street},
            city={Toronto},
            postcode={ON M3J 1P3},
            state={Ontario},
            country={Canada}}

\affiliation[6]{organization={University of Groningen},
            addressline={Nijenborgh 4},
            city={Groningen},
            postcode={9747 AG},
            country={Netherlands}}

\begin{abstract}
Detecting disinformation that blends manipulated text and images has become increasingly challenging, as AI tools make synthetic content easy to generate and disseminate. While most existing AI-safety benchmarks focus on single-modality misinformation (i.e., false content shared without intent to deceive), intentional multimodal disinformation, such as propaganda or conspiracy theories that imitate credible news; remains largely unaddressed.
In this work, we introduce the \textbf{V}ision-\textbf{L}anguage \textbf{D}isinformation Detection \textbf{Bench}mark (\textsf{\textbf{\textsc{VLDBench}}}), the first large-scale resource supporting both unimodal (text-only) and multimodal (text + image) disinformation detection. \textsf{\textbf{\textsc{VLDBench}}} comprises approximately 62,000 labeled text–image pairs across 13 categories, curated from 58 news outlets. Using a semi-automated pipeline followed by expert review, 22 domain experts invested over 500 hours to produce high-quality annotations with substantial inter-annotator agreement.
Evaluation of state-of-the-art LLMs and VLMs on \textsf{\textbf{\textsc{VLDBench}}} shows that adding visual cues improves detection accuracy, with gains ranging from ~5 points for strong baselines (e.g., LLaMA-3.2-11B-Vision 74.82\% vs.\ LLaMA-3.2-1B-Instruct 70.29\%) to ~25–30 points for smaller families (e.g., LLaVA-v1.5-Vicuna7B 72.32\% vs.\ Vicuna-7B-v1.5 55.21\%), reflecting complementary evidence from images (e.g., meme-like visuals, image–text consistency) that text alone cannot capture. We provide data and code for evaluation, fine-tuning and robustness tests to support disinformation analysis. Developed in alignment with the AI Goverance frameworks (MIT AI Risk Repository), \textsf{\textbf{\textsc{VLDBench}}} offers a principled foundation for advancing trustworthy disinformation detection in multimodal media.
\begin{center}
\begin{tabular}{cll}
\faGlobe & \textbf{Project:} & {\small\url{https://vectorinstitute.github.io/VLDBench/}} \\
\faDatabase & \textbf{Data:} & {\small\href{https://huggingface.co/datasets/vector-institute/VLDBench}{\nolinkurl{https://huggingface.co/datasets/vector-institute/VLDBench}}} \\
\faGithub & \textbf{Code:} & {\small\url{https://github.com/VectorInstitute/VLDBench}}
\end{tabular}
\end{center}

\end{abstract}




\end{frontmatter}

\section{Introduction}
The global proliferation of false or misleading content has reached alarming levels: approximately 60\% of people worldwide report encountering false narratives on digital platforms \footnote{\href{https://redline.digital/fake-news-statistics/}{Fake news statistics (2024)}}
, and 94\% of journalists view fabricated news as a significant threat to public trust \footnote{\href{https://www.pewresearch.org/journalism/2022/06/14/journalists-highly-concerned-about-misinformation-future-of-press-freedoms/}{Pew Research Center (2022) Report on Journalism}}. The World Economic Forum \footnote{\href{https://www3.weforum.org/docs/WEF_The_Global_Risks_Report_2024.pdf}{World Economic Forum, \textit{Global Risks Report 2024}}}
 identifies \textbf{misinformation} (unintentionally shared false information) and \textbf{disinformation} (deliberately deceptive content) among the leading global risks for 2024. Distinguishing between these two AI risks is crucial: while misinformation spreads without malicious intent \cite{santos2021misinformation}, disinformation is strategically weaponized, exploiting digital ambiguity to erode societal trust \footnote{\href{https://commonslibrary.org/disinformation-and-7-common-forms-of-information-disorder/}{7 Common Forms of Information Disorder}}. In this work, we restrict our scope to disinformation manifested in multimodal (text–image) contexts.

\begin{figure}[h]
    \centering
    \includegraphics[width=0.88\linewidth]{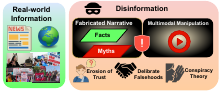}
    \caption{
      Schematic illustration of the disinformation flow: fabricated narratives and multimodal manipulation distort real-world information, producing myths and deliberate falsehoods and ultimately eroding public trust.
    }
    \label{fig:disinfo-flow}
\end{figure}

Generative AI (GenAI) has recently played a paradoxical role, as it can both generate and detect disinformation \cite{bontcheva2024generative}.  For instance, text-to-image models can produce convincing but fabricated visuals of public figures at events that never occurred, while the same models can also be leveraged to flag image–text inconsistencies, such as detecting when an old protest photo is falsely paired with a recent political headline. Although recent advances have improved AI-driven detection methods, addressing disinformation in real-world news (as illustrated in Figure~\ref{fig:disinfo-flow}) remains challenging. While most research continues to focus on text-only cases, multimodal disinformation: such as repurposed images used to reinforce fabricated narratives; is still significantly underexplored, posing complex technical challenges and serious societal implications.

Governments and technology companies are increasingly addressing disinformation through emerging regulatory frameworks. Policies such as the EU Action Plan \cite{EUActionPlan2018}, the U.S. Digital Fairness Act\footnote{\href{https://www.nysenate.gov/legislation/bills/2023/A3308}{New York State Senate — Bill A3308 (2023)}}, and Canada's AI initiatives \cite{canada_aia} (summarized in Table~\ref{tab:disinformation_regulations}) emphasize transparency, accountability, and responsible AI deployment.
In parallel, industry leaders—including Google\footnote{\href{https://ai.google/principles/}{Google AI Principles}}, Meta\footnote{\href{https://about.facebook.com/meta/}{Meta Principles}}, Microsoft\footnote{\href{https://www.microsoft.com/en-us/research/project/ai-fairness-checklist/}{Microsoft AI Fairness Checklist}}, and Anthropic \cite{anthropic2023core}, have launched technical countermeasures through collaborative initiatives, outlined in Table~\ref{tab:bodies}. 
The MIT AI Risk Repository \cite{slattery_2024ai}, which synthesizes over 1,600 risks across 65 AI frameworks, explicitly identifies disinformation as a systemic threat requiring robust governance mechanisms (see Appendix~\ref{app:airr}). However, despite these efforts, governance frameworks and industry pledges remain largely principle-level; they lack practical benchmarking resources that operationalize and audit performance on \emph{multimodal} disinformation detection. 
This gap motivates VLDBench, a governance-aligned, human-verified benchmark for evaluating text–image disinformation.

Existing safety benchmarks for large language models (LLMs), such as SafetyBench \cite{zhang2023safetybench} and AISB \cite{vidgen2024introducingv05aisafety}; primarily target general harms (e.g., toxicity, bias). Large Vision–language models (LVLMs) safety benchmarks, including MM-SafetyBench \cite{liu2025mm}, MultiTrust \cite{zhang2024multitrust} (social bias), VLGuard \cite{zong2024safety}, ALM-Bench \cite{vayani2024all} (cultural bias), and Rainbow Teaming \cite{samvelyan2024rainbowteaming}, emphasize robustness and alignment but  the specific threat of disinformation is not under specific topics. Some benchmarks, such as Misinformer \cite{papado2023misinformer}, Vision–Language Deception Detection \cite{Qi_2024_CVPR}, and deepfake-oriented resources \cite{li2024towards}, incorporate multimodal elements, yet intentional multimodal disinformation (e.g., propaganda or health scams pairing misleading text with incongruent or manipulated imagery) remains out of scope.

GenAI is reshaping how people access and produce news, with text-image systems able to fabricate persuasive content at scale. Policymakers now warn not only about user deception but also about damage to the \emph{digital commons}: The European Commission’s JRC Generative AI Outlook (2025) \footnote{\href{https://op.europa.eu/en/publication-detail/-/publication/9f7e0b86-477c-11f0-85ba-01aa75ed71a1/language-en}{The European Commission’s JRC Generative AI Outlook (2025)}} cautions that ``AI generated errors and disinformation can pollute open knowledge repositories (e.g., Wikipedia), creating costly verification burdens". Therefore, in this paper we introduce \textsf{\textbf{\textsc{VLDBench}}}, a multimodal benchmark for news style text-image disinformation detection, including robustness stress tests and governance- aware reporting. Our goal is to offer a rigorous, open resource to measure and mitigate risks before synthetic content contaminates shared knowledge ecosystems.

 To our knowledge, \textsf{\textbf{\textsc{VLDBench}}} is first comprehensive, human-verified benchmark aligned with emerging AI governance standards, supporting accountability, transparency, and robustness in disinformation detection across both unimodal (text-only LLMs) and multimodal (text-image) systems. Table~\ref{tab:comparison} compares our benchmark with prior efforts, showing that existing studies primarily target misinformation and seldom address disinformation.

 \begin{table}[t]
\centering
\caption{Comparison of \textbf{\textsc{VLDBench}} with contemporary datasets. The \textit{\textbf{Annotation}} (\textcolor{blue}{\faUser}) means Manual and (\textcolor{blue}{\faUser},\textcolor{purple}{\faCog}) indicates Hybrid (Human+AI). \textit{\textbf{Access}?} (\textcolor{mygreen}{\ding{51}}) refers to open-source and (\textcolor{myred}{\faExclamationTriangle}) indicates Request required. The \textit{\textbf{Real}?} is defined as (\textcolor{mygreen}{\ding{51}}) if it's real-world data and (\textcolor{red}{\ding{55}}) if it's a Synthetic data. \textsuperscript{*}\textbf{Multiple} includes \textit{Politics, National, Business \& Finance, International, Local/Regional, Entertainment, Opinion/Editorial, Health, Other, Sports, Technology, Weather \& Environment, Science}.}
\footnotesize
\resizebox{\textwidth}{!}{
\begin{tabular}{|l|l|c|c|c|c|c|}
\hline
\textbf{Dataset} & \textbf{Primary Task} & \textbf{Multimodal} & \textbf{Real?}& \textbf{Category} & \textbf{Annotation type}& \textbf{Access?} \\
\hline
LIAR \cite{wang-2017-liar} & Misinfo. (Text) & \textcolor{red}{\ding{55}} & \textcolor{mygreen}{\ding{51}} & Politics & \textcolor{blue}{\faUser} & \textcolor{mygreen}{\ding{51}} \\
\hline
FakeNewsNet \cite{shu2017fake} & Misinfo. (Text+Image) & \textcolor{mygreen}{\ding{51}} & \textcolor{mygreen}{\ding{51}} & Politics, Social & \textcolor{blue}{\faUser} & \textcolor{myred}{\faExclamationTriangle} \\
\hline
BuzzFace \cite{Santia_Williams_2018} & Misinfo. (Text) & \textcolor{red}{\ding{55}} & \textcolor{mygreen}{\ding{51}} & Politics & \textcolor{blue}{\faUser} & \textcolor{mygreen}{\ding{51}} \\
\hline
FEVER \cite{thorne2018feverlargescaledatasetfact} & Misinfo. (Text) & \textcolor{red}{\ding{55}} & \textcolor{mygreen}{\ding{51}} & Social Media & \textcolor{blue}{\faUser} & \textcolor{myred}{\faExclamationTriangle} \\
\hline
RealNews \cite{zellers2019defending} & Misinfo. (Text) & \textcolor{red}{\ding{55}} & \textcolor{mygreen}{\ding{51}} & General News & \textcolor{blue}{\faUser} & \textcolor{myred}{\faExclamationTriangle} \\
\hline
Nela-GT \cite{gruppi2021nela} & Misinfo. (Text) & \textcolor{red}{\ding{55}} & \textcolor{mygreen}{\ding{51}} & Politics, COVID-19 & \textcolor{blue}{\faUser},\textcolor{purple}{\faCog} & \textcolor{myred}{\faExclamationTriangle} \\
\hline
Fakeddit \cite{nakamura-etal-2020-fakeddit} & Misinfo. (Text+Image) & \textcolor{mygreen}{\ding{51}} & \textcolor{mygreen}{\ding{51}} & Social Media & \textcolor{blue}{\faUser},\textcolor{purple}{\faCog} & \textcolor{mygreen}{\ding{51}} \\
\hline
NewsBag \cite{NewsBag2020dataset} & Misinfo. (Text+Image) & \textcolor{mygreen}{\ding{51}} & \textcolor{mygreen}{\ding{51}} & Politics & \textcolor{blue}{\faUser},\textcolor{purple}{\faCog} & \textcolor{mygreen}{\ding{51}} \\
\hline
MuMiN \cite{mumin2022dataset} & Misinfo. (Text+Image) & \textcolor{mygreen}{\ding{51}} & \textcolor{mygreen}{\ding{51}} & Social Media & \textcolor{blue}{\faUser} & \textcolor{myred}{\faExclamationTriangle} \\
\hline
DGM\textsuperscript{4} \cite{shao2024dgm4++} & Disinfo. (Deep Fake Images) & \textcolor{mygreen}{\ding{51}} & \textcolor{red}{\ding{55}} & General News & \textcolor{blue}{\faUser},\textcolor{purple}{\faCog} & \textcolor{mygreen}{\ding{51}} \\
\hline
\rowcolor{lightgray!30}
\textbf{\textsc{VLDBench}} & \textbf{Disinfo. (Text+Image)} & \textcolor{mygreen}{\ding{51}} & \textcolor{mygreen}{\ding{51}} & \textbf{Multiple\textsuperscript{*}} & \textcolor{blue}{\faUser},\textcolor{purple}{\faCog} & \textcolor{mygreen}{\ding{51}} \\
\hline
\end{tabular}
}

\label{tab:comparison}
\end{table}

\begin{wrapfigure}{r}{0.5\textwidth}
    \centering
    \vspace{-10pt}
    \includegraphics[width=\linewidth]{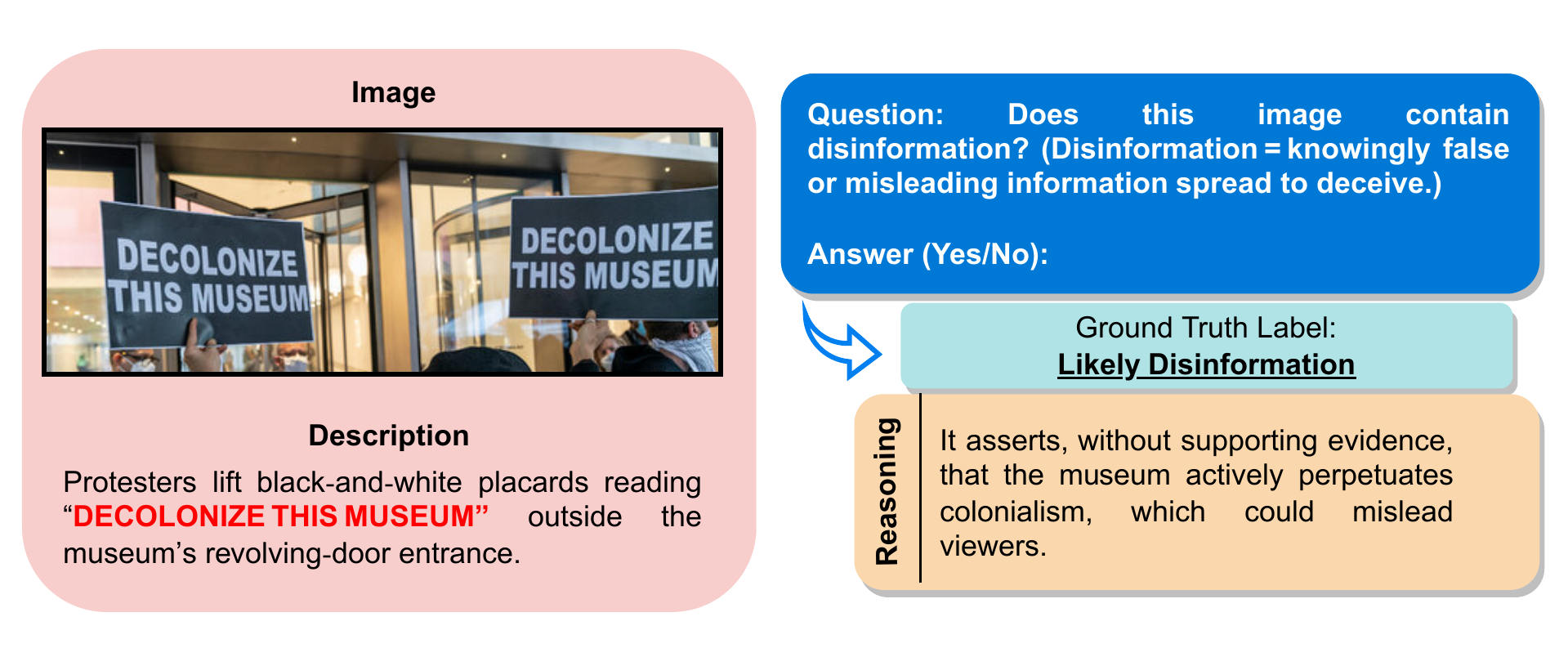}
   \caption{\textbf{Disinformation example.} This instance spreads likely disinformation by generating political hype without factual basis.}
    \label{fig:example}
    \vspace{-10pt}
\end{wrapfigure}

\paragraph{Contributions} 
The main contributions of this work are as follows:  

\begin{itemize}[itemsep=0pt]
    \item \textbf{VLDBench benchmark.} We present \textsf{\textbf{\textsc{VLDBench}}}, the largest human-verified benchmark for multimodal disinformation detection. It comprises 31,339 unique news articles with paired images, curated from 58 outlets across 13 topical categories, yielding a total of 62,678 fully verified instances (31,339 text-only and 31,339 text–image pairs). All data and code are released to support open research and reproducibility.  

    \item \textbf{Task coverage.} \textsf{\textbf{\textsc{VLDBench}}} supports two complementary evaluation settings: (i) binary classification of text or text–image pairs, and (ii) open-ended multimodal reasoning. This dual coverage enables both traditional supervised evaluation and generative reasoning analysis (see Figure~\ref{fig:example}).  

    \item \textbf{Semi-autonomous annotation pipeline.} We introduce a scalable annotation workflow combining LLM-assisted labeling (GPT-4o) with rigorous expert verification on all samples to ensure reliability and efficiency.  

    \item \textbf{Expert annotation quality.} Twenty-two domain experts invested over 500 hours in verification, achieving substantial agreement (Cohen’s $\kappa=0.78$), resulting in one of the highest-quality disinformation benchmarks to date.  

    \item \textbf{Comprehensive benchmarking.} We benchmark state-of-the-art multimodal LVLMs and text-only  LLMs on \textsf{\textbf{\textsc{VLDBench}}}, uncovering systematic performance gaps, modality-specific vulnerabilities, and model-specific failures with direct implications for responsible AI governance and risk monitoring.  
\end{itemize}

Our evaluation reveals that LVLMs consistently outperform text-only LLMs in disinformation detection. For example, the top-performing VLM, LLaMA-3.2-11B-Vision, achieves 74.82\% accuracy on \textsf{\textbf{\textsc{VLDBench}}}, exceeding the best text-only LLM (LLaMA-3.2-1B-Instruct, 70.29\%) by 4.53 percentage points (Section~\ref{sec:results}).  ThESE results affirms that disinformation extends beyond textual content and thus multimodality is paramount for detection.
We also find that while both LLMs and lVLMs exhibit robustness to minor perturbations in single modalities (e.g., image noise or textual paraphrasing), their performance degrades sharply under cross-modal attacks that combine distortions in both text and image inputs. 
The data, code, and findings from this work can support future research on disinformation tactics, and we invite researchers and developers to build upon them.

\section{Related Work}

\subsection{Disinformation as a Global Threat and AI Goverance Bodies}
Disinformation is now widely recognised as a systemic risk in AI governance.  At the supranational level, the EU’s \emph{Digital Services Act} (DSA) \cite{DigitalServicesAct} imposes obligations on online intermediaries and very large platforms to prevent illegal and harmful activities and mitigate systemic risks such as manipulation and disinformation.  The strengthened \emph{Code of Practice on Disinformation} , which was formally incorporated into the DSA in 2025, aims to combat disinformation while upholding freedom of expression and enhancing transparency; its commitments are now auditable for designated platforms under the DSA.  The UK’s \emph{Online Safety Act 2023} \cite{UK_Online_Safety_Act_2023} takes a proportionate approach to mis- and disinformation: it requires services to remove illegal or state-sponsored disinformation once aware of it, mandates additional protections when children could be exposed, and preserves users’ freedom of expression.

In Canada, the \emph{Online News Act (Bill C‑18)} \cite{Canada_Online_News_Act_2023} does not regulate disinformation directly but seeks to sustain an informed citizenry by obligating dominant platforms to compensate news publishers for their content.  The European Union’s \emph{AI Act} \cite{EU_AI_Act_2024}, the world’s first comprehensive AI law adopts a risk-based framework that bans certain “unacceptable risk” AI practices, including harmful AI-based manipulation and deception.  These legislative instruments, summarised in Table\ref{tab:disinformation_regulations}, signal a growing consensus that effective disinformation governance requires combining policy obligations with technical safeguards, while respecting fundamental rights.

\begin{table}[t]
\centering
\caption{Global regulations and acts addressing disinformation.}
\label{tab:disinformation_regulations}
\small
\setlength{\tabcolsep}{6pt}\renewcommand{\arraystretch}{1.15}
\resizebox{\textwidth}{!}{
\begin{tabular}{@{}>{\raggedright\arraybackslash}p{5.2cm} >{\raggedright\arraybackslash}p{11.6cm}@{}}
\toprule
\textbf{Regulation / Act} & \textbf{Focus and relevance} \\
\midrule
EU Code of Practice on Disinformation (2018; strengthened 2022) &
Defines “verifiably false or misleading information,” and commits signatories to demonetisation, fact-checking access, transparency reporting, and researcher data access. \\
European Democracy Action Plan (EU) &
Whole-of-government policy to protect elections and media freedom and counter disinformation; it informed and supported the strengthened Code. \\
Digital Services Act (EU) &
Requires VLOPs/VLOSEs to assess systemic risks (Art.~34, incl.\ disinformation) and adopt mitigation measures (Art.~35), with audits and Commission enforcement. \\
European Union \emph{AI Act} (2024) &
Transparency duties (e.g., labelling certain synthetic/deepfake content and disclosure requirements for some AI interactions) under a risk-based regime.
\cite{EUCodeDisinfo,EUDemocracyActionPlan,DigitalServicesAct,EUAIAct2024} \\
\midrule
Countering Foreign Propaganda and Disinformation Act (US, 2016) &
Establishes/strengthens the Global Engagement Center to coordinate counter-disinformation efforts against foreign state actors. \cite{CounteringForeignPropaganda} \\
\textit{(Proposed)} Malicious Deep Fake Prohibition Act (US) \& related actions &
Federal bill not enacted; related actions include FCC treatment of AI-generated voices in robocalls as “artificial,” enabling enforcement against deceptive calls.
\cite{MaliciousDeepFakeProhibition,FCC_AIRobocalls_2024} \\
Online Safety Act (UK, 2023) &
Duties on platforms to reduce illegal harms and manage risks; Ofcom enforcement with penalties. \cite{OnlineSafetyActUK} \\
Deep Synthesis Provisions (China, 2022/2023) &
Labelling requirements for synthetic media and obligations on deep synthesis service providers to prevent misuse. \cite{ChinaDeepSynthesisProvision} \\
Australian Code of Practice on Disinformation and Misinformation (DIGI) &
Industry code with transparency reporting and complaints mechanism; complements prospective regulator powers. \cite{DIGI2022Code} \\
Singapore POFMA &
Correction directions and takedown orders for false statements of fact in the public interest. \cite{SingaporePOFMA2019} \\
Online News Act (Canada, 2023) &
Regulates digital news intermediaries to support a sustainable news ecosystem (indirectly affecting disinformation incentives). \\
Disinformation Guidebook (Canada, 2024) &
Guidance for public servants on identifying and responding to disinformation and building resilience.
\cite{OnlineNewsAct2023,govCanada_disinformation} \\
\bottomrule
\end{tabular}
}
\end{table}

\begin{table}[h]
\centering
\footnotesize
\setlength{\tabcolsep}{6pt}
\renewcommand{\arraystretch}{1.2}
\caption{Key Industry Initiatives for Responsible AI and Disinformation Mitigation}
\label{tab:bodies}
\begin{tabularx}{\textwidth}{@{}>{\raggedright\arraybackslash}p{3cm} >{\raggedright\arraybackslash}p{4cm} >{\raggedright\arraybackslash}X@{}}
\toprule
\textbf{Initiative} & \textbf{Members / Founders} & \textbf{Objective / Focus} \\
\midrule
Frontier Model Forum (FMF) 
 & Founders: Anthropic, Google, Microsoft, OpenAI; later joined by Amazon and Meta (2024). & Advance frontier AI safety research, best-practice standards, and information-sharing; grants and technical briefs in 2024–2025.  \\
\addlinespace[2pt]
Partnership on AI (PAI) - Synthetic Media Framework \cite{pai2024} & Broad multi-stakeholder consortium (e.g., Google, Microsoft, Meta, Amazon, IBM, Adobe, Apple, NGOs, academia). & Responsible practices for synthetic media; 2024 case studies and policy recommendations on disclosure/labelling.  \\
\addlinespace[2pt]
C2PA (Coalition for Content Provenance and Authenticity)  \cite{c2pa2024}  & Adobe, Microsoft, Intel, Arm, Truepic (and many others) & Open technical standard for content credentials/provenance (spec v2.2+), enabling tamper-evident media and deepfake attribution.\\
\addlinespace[2pt]
Tech Accord on AI and Elections (2024)  \cite{techaccord2024} & Adobe, Amazon, Google, IBM, Meta, Microsoft, OpenAI, TikTok, X, and others & Joint commitments to detect, label, and curb deceptive AI election content; rapid response and user-education pledges. \\
\bottomrule
\end{tabularx}
\end{table}

\subsection{Industry-led Initiatives}
In addition to regulatory acts, several industry-led bodies play a complementary role in disinformation mitigation.  The \emph{Frontier Model Forum} \footnote{\href{https://blogs.microsoft.com/on-the-issues/2023/07/26/anthropic-google-microsoft-openai-launch-frontier-model-forum/}{Microsoft blog post on Frontier Model Forum}}
, a non-profit supported by Anthropic, Google, Microsoft and OpenAI, draws on member expertise to ensure that “frontier” AI systems remain safe and secure; its mandates include identifying best practices, advancing independent research and facilitating information sharing across government, academia and industry.  The \emph{Partnership on AI} (PAI) \cite{pai2024} is a multi‑stakeholder non-profit that now has over 100 partners worldwide; its programme on “AI \& Media Integrity” seeks to address misinformation and ensure that vulnerable communities are not disproportionately harmed by AI-driven content.  The \emph{Coalition for Content Provenance and Authenticity} (C2PA) \cite{c2pa2024}, hosted by the Linux Foundation, provides an open technical standard for “Content Credentials” that functions like a “nutrition label” for digital media, enabling publishers and consumers to trace the origin and edits of content and thereby promote transparency.  Finally, the \emph{Tech Accord to Combat Deceptive Use of AI in Elections} \footnote{\href{https://blogs.microsoft.com/on-the-issues/2024/02/16/ai-deepfakes-elections-munich-tech-accord/}{AI, Deepfakes and Elections – Munich Tech Accord}}
 is a 2024 pledge signed by more than twenty major technology companies,including Adobe, Amazon, Google, IBM, Meta, Microsoft, OpenAI, Snap and TikTok, to develop tools that detect and counter harmful AI-generated election content; the accord emphasises tracking the origin of deceptive media and raising public awareness.  Together with statutory AI acts (such as the EU AI Act) that prohibit certain manipulative uses of AI \cite{EUCodeDisinfo, EuropeanCommission2020}, these initiatives illustrate how governance frameworks and industry commitments converge on the dual imperative of mitigating harmful content while safeguarding free speech.

\subsection{Disinformation Detection Methodologies}
Disinformation deliberately false or misleading content, often leverages partisan framing, visuals and contextual cues to manipulate public opinion \cite{CPA2024Handbook}.  Early detection approaches relied on linguistic heuristics and rule-based systems \cite{rubin2016fake}, which generalised poorly.  Classical machine-learning pipelines using RNNs, CNNs and variational autoencoders \cite{bahad2019fake,khattar2019mvae} improved adaptability but lacked deeper semantic understanding.  With the rise of multimodal content, researchers developed architectures such as CARMN \cite{song2021multimodal}, MCNN \cite{segura2022multimodal} and MCAN \cite{wu2021multimodal} that fuse text–image representations.  More recently, large language models (LLMs) like ChatGPT offer greater contextual reasoning \cite{chen2023combating}, but they also lower the barrier to producing convincing disinformation, creating a dual-use dilemma.  LVLMs extend detection by aligning visual and textual modalities \cite{papado2023misinformer,Qi_2024_CVPR,xuan2024lemma}.  Knowledge-augmented systems such as FakeNewsGPT4 \cite{liu2024fakenewsgpt4} and LEMMA \cite{xuan2024lemma} enhance robustness but still struggle with context sensitivity and subjective interpretation.  Separate lines of work probe demographic bias in LVLMs, e.g., assigning higher confidence to male versus female doctors \cite{wan2025evaluating} and create evaluation suites (VLBiasBench \cite{zhang2024vlbiasbench}, MultiTrust \cite{zhang2024multitrust}, MM‑SafetyBench \cite{liu2025mm}) to audit fairness and robustness.  These studies highlight the need to address hallucination, bias and safety risks alongside classification performance.

We build on extensive prior work and view our contribution as complementary to these foundations. Text-only corpora (e.g., LIAR \cite{wang-2017-liar}, FEVER \cite{thorne2018feverlargescaledatasetfact}) are pioneering resources in this line of research and have catalyzed methodological advances; by design, however, they do not capture the cross-modal inconsistencies that adversaries increasingly exploit. Social-context repositories (e.g., FakeNewsNet \cite{shu_fakenewsnet_2020} with PolitiFact/GossipCop) enrich analysis with propagation and engagement signals, yet most benchmarks in this family still emphasize unimodal text. Multimodal collections (e.g., Fakeddit \cite{nakamura-etal-2020-fakeddit}, Factify/Factify-2 \cite{suryavardan2023factify2multimodalfake}, MuMiN \cite{mumin2022dataset}) advance text–image evaluation at scale; many necessarily employ distant supervision (heuristic labels, weak source tags, or other proxy signals) to enable coverage, which can introduce ambiguity and complicate fine-grained robustness audits \cite{farooq2025evaluating}.

Out-of-context and image-forensics suites (e.g., News out-of-context \cite{luo2021newsclippings}, FaceForensics++ \cite{roessler2019faceforensicspp}, DFDC \cite{dolhansky2020dfdc}, Celeb-DF \cite{li2020celeb}) are invaluable for studying synthetic media and visual manipulation, though their objectives differ from newsroom-style disinformation and they typically do not pair verified article text with contemporaneous images. Topic-specific credibility sets from the COVID-19 period (e.g., ReCOVery \cite{zhou2020recovery}, CoAID \cite{cui2020coaid}) provided timely coverage of a critical domain; their topical scope and collection window, however, reflect conditions at the time of curation. As summarized in Table~\ref{tab:comparison}, differences in objective, modality coverage, and collection period leave opportunities for resources aligned with today’s LLM/VLM evaluation practices and emerging transparency expectations. 

\textsf{\textbf{\textsc{VLDBench}}} is designed to address this opportunity: it provides temporally recent, human-verified article–image pairs across diverse news domains, explicit \emph{disinformation} labels (beyond generic misinformation), robustness stress-tests, and traceable annotation linked to accountability and transparency principles. Next, we present our methodology.

\begin{figure}[t]
\centering
\setlength{\tabcolsep}{6pt}
\renewcommand{\arraystretch}{1.05}
\begin{tabular}{@{}p{0.48\linewidth} p{0.48\linewidth}@{}}

\parbox[t]{\linewidth}{%
  \includegraphics[width=\linewidth,height=0.28\textheight,angle=180]{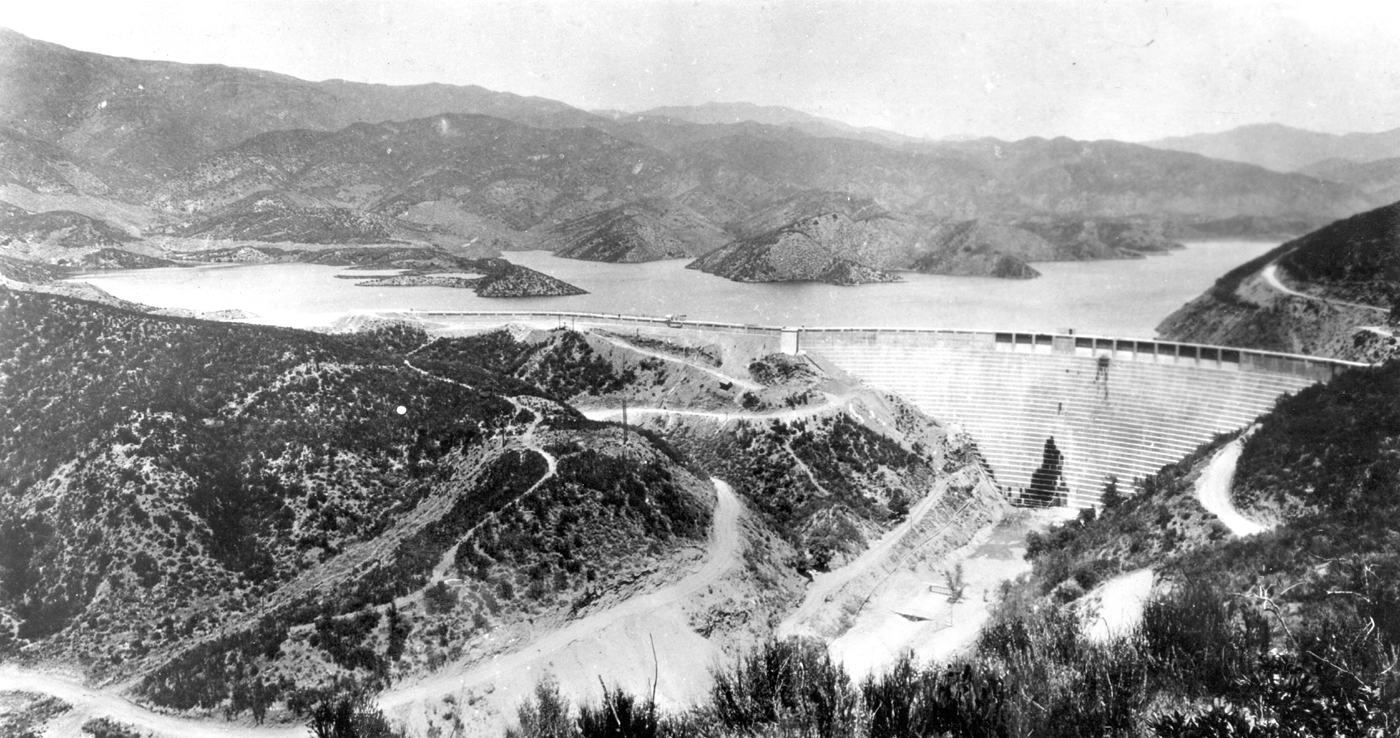}\\[2pt]
  \footnotesize\emph{Representative Example}

}
&
\parbox[t]{\linewidth}{\small
  \textbf{Disinformation Example (text–image)}\\[2pt]
  \textbf{Headline:} City declares \issue{\emph{new}} water crisis after dam \issue{“collapse”}; \issue{boil all tap water immediately}.\\[2pt]
  \textbf{Text:} “Officials confirm the main dam failed last night. \issue{Share this now} so your family stays safe”.\\
  \textbf{Rationale:} This pair is labelled \emph{Likely Disinformation} because it combines a verifiable \textbf{false/manipulated element} with persuasive use of that same element. 
  The image is misdated/misattributed because an older incident presented as \issue{\emph{new}} and no current failure appears in official notices; the urgent language (\issue{“boil all tap water,” “share this now”}) is deployed to make readers accept and act on the false claim of a \issue{“collapse”}. 
  Style or tone on its own would not qualify.
}
\end{tabular}
\caption{Representative example: left = image; right = concise rule and evidence.}
\label{fig:disinfo-example}

\end{figure}

\section{\underline{V}ision \underline{L}anguage \underline{D}isinformation Detection \underline{Bench}mark} 
\label{method}  
\subsection{Dataset overview}
\begin{figure}[t]
    \centering
    \includegraphics[width=\linewidth]{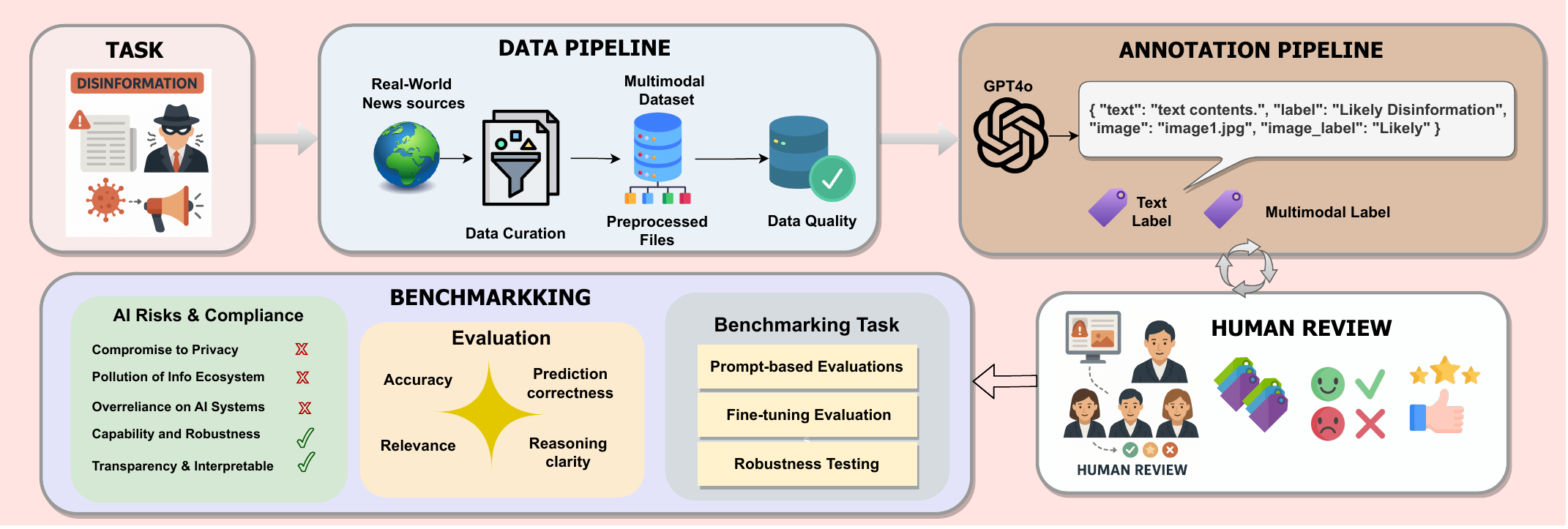}
    \caption{\textsf{\textbf{\textsc{VLDBench}}} Framework:  The system comprises the stages: (1) Define Task : formalizing the detection objective; (2) Data Pipeline : curating and preprocessing real-world multimodal news content; (3) Annotation Pipeline : generating labels via LLM-assisted for scale and; (4) Human Review : validating annotations through expert oversight; (5) Benchmarking : evaluating models for accuracy, reasoning, and risk mitigation across prompting, fine-tuning, and robustness scenarios.}
    \label{fig:main}
\end{figure}

\textsf{\textbf{\textsc{VLDBench}}} (Figure~\ref{fig:main}) is a comprehensive benchmark for multimodal disinformation detection. It contains 31{,}339 unique news articles, each paired 1:1 with its accompanying image, yielding 62{,}678 instances in total (31{,}339 text-only and 31{,}339 text–image pairs). All labels are human-verified to ensure reliability across modalities. The corpus is curated from 58 news sources, including the \emph{Financial Times}, CNN, \emph{The New York Times}, and \emph{The Wall Street Journal} (see Figure~\ref{fig:news_sources_distribution}). It spans 13 categories—\textit{National, Business \& Finance, International, Entertainment, Local/Regional, Opinion/Editorial, Health, Sports, Politics, Weather \& Environment, Technology, Science, Other}—as shown in Figure~\ref{fig:news_categories}. Further descriptive statistics appear in Section~\ref{analysis}.

\subsection{Task Definition}

\textbf{Disinformation detection.}
We define \emph{disinformation} as misleading or manipulated information, textual or visual, intentionally created or disseminated to deceive, harm, or influence individuals, groups, or public opinion. This follows social-science literature \cite{benkler2018network} and international guidance, emphasizing \emph{intent} rather than mere factual inaccuracy. We distinguish \emph{misinformation} (false or misleading content shared \emph{without} intent to deceive) from \emph{disinformation} (content that is \emph{intentionally} false or manipulated to persuade, harm, or influence public opinion) \cite{santos2021misinformation}. Our focus is on disinformation.

An item is labeled \emph{Likely Disinformation} only when a verifiably false claim or manipulated element is present \emph{and} that same element is used to persuade the reader; otherwise it is labeled \emph{Unlikely Disinformation} (Figure~\ref{fig:disinfo-example}). We formulate detection as binary classification over (i) standalone text or (ii) text–image news pairs. Given input $x$ (article text, or text–image pair), the model predicts $y \in \{\text{Likely Disinformation}, \text{Unlikely Disinformation}\}$ and provides a brief rationale.

\subsection{Data Curation Pipeline}  
\textbf{Dataset Collection:}  
We collected news articles via Google RSS feeds from a broad spectrum of outlets (listed in Table~\ref{tab:sources}), in accordance with Google’s Terms of Service \footnote{\href{https://policies.google.com/terms}{Google Terms of Service}}
. The collection period spanned from \textbf{May 6, 2023} to \textbf{September 6, 2024}, covering major global events. Each record includes the full article text and the primary image, capturing a diverse range of topics. All data handling followed established ethical guidelines on intellectual property rights and privacy protection \footnote{\href{https://uwaterloo.ca/research/office-research-ethics/research-human-participants/pre-submission-and-training/human-research-guidelines-policies-and-resources/does-my-data-collection-activity-require-ethics-review}{UWaterloo Ethics Review Guidance}}
.

\textbf{Pre-processing and Filtration.}  
We applied structured filtering to preserve data integrity. Entries were removed if they contained incomplete text, missing or low-resolution images, duplicates, or media-centric URLs (e.g., \texttt{/video}, \texttt{/gallery}). Articles with fewer than 20 sentences were discarded to ensure sufficient textual depth. For each retained article, the lead image was designated as the visual counterpart. We then audited a stratified sample to confirm that the RSS API returned valid and internally consistent records. These procedures yielded \textbf{31{,}339} unique curated text–image news articles, which were subsequently routed to the annotation pipeline.

\subsection{Data Annotation Pipeline}
We used GPT-4o \cite{gpt4o} as an LLM annotator to enable scalable labeling, leveraging its multimodal reasoning capabilities and consistent performance across image–text pairs \cite{kim2024meganno+}. (Other LLMs can be substituted without changing the protocol.) 

\begin{wrapfigure}{r}{0.48\textwidth}
    \vspace{-5pt} 
    \centering
    \includegraphics[width=0.4\textwidth]{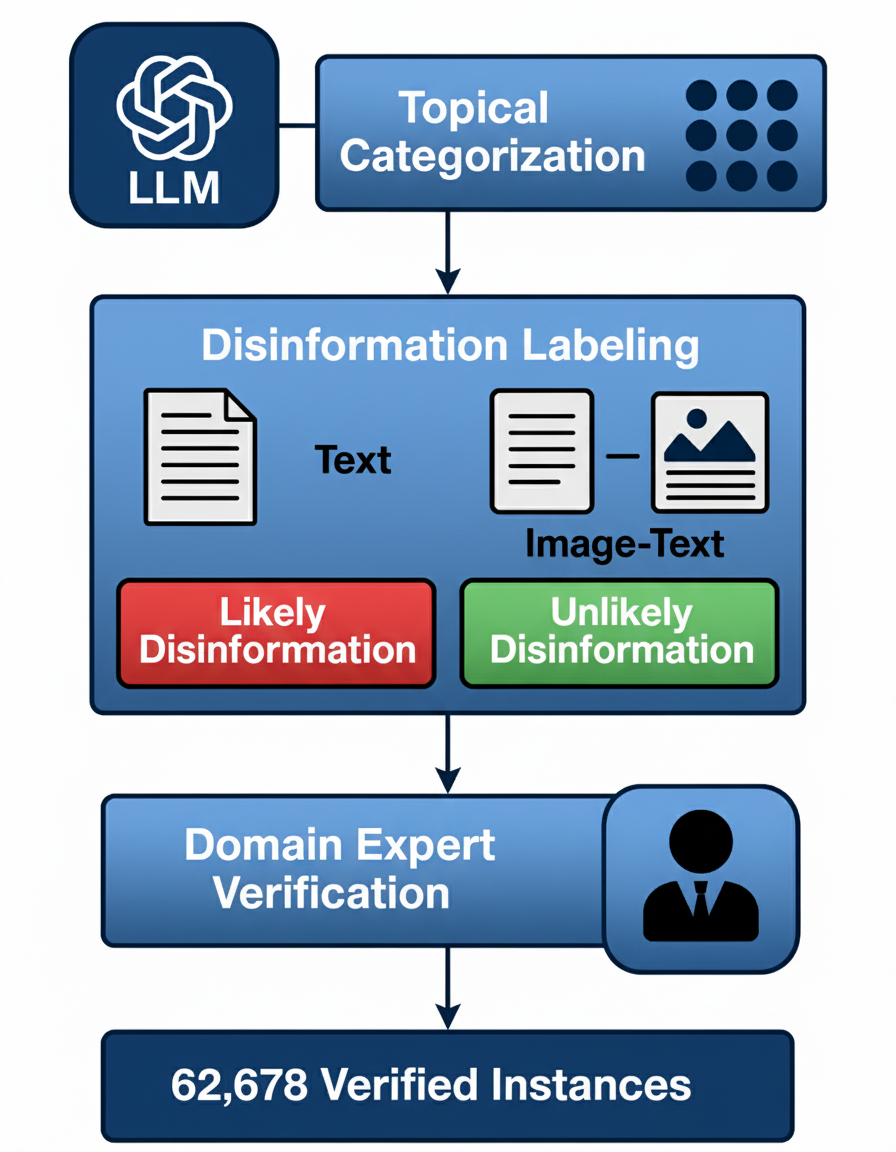}
    \caption{Annotation pipeline used in \textsf{\textbf{\textsc{VLDBench}}}.}
    \label{fig:annotation_pipeline}
    \vspace{-2pt}
\end{wrapfigure}%
\textbf{Topical Categorization.}
Following preprocessing, GPT-4o categorized each image–text pair into one of 13 news categories: \textit{National, Business \& Finance, International, Entertainment, 
Local/Regional, Opinion/Editorial, Health, Sports, Politics, Weather \& Environment, Technology and Science} (as shown in Figure~\ref{fig:news_categories}). The category design was informed by the AllSides media bias chart \cite{allsides_mediabiaschart} and the Media Cloud taxonomy \footnote{\href{https://www.mediacloud.org/}{Media Cloud}}
.\\
\textbf{Disinformation Labeling.}
GPT-4o then evaluated every article, producing a binary label (\texttt{Likely Disinformation} or \texttt{Unlikely Disinformation}) and a concise rationale. To control stochastic variance, each instance was assessed three times; majority vote determined the final label. For the unimodal set, GPT-4o was prompted exclusively with the article text (no image), followed by human verification. For the multimodal set, GPT-4o received both the text and its paired image to ensure labels reflect joint semantics rather than a single modality. The pipeline yielded \textbf{62{,}678} fully verified instances: \textbf{31{,}339} text-only articles and \textbf{31{,}339} text–image pairs (Table~\ref{tab:dataset_statistics}).

\subsection{Domain Expert Verification}
To our knowledge, \textsf{\textbf{\textsc{VLDBench}}} is the largest human-verified \emph{disinformation} benchmark to date, reflecting \textbf{500+ hours} of expert review. A team of 22 domain experts from diverse disciplines systematically reviewed all 62k samples, including GPT-4o labels and rationales, assessing accuracy, consistency, and alignment with human judgment. Two subject-matter experts in computer science and linguistics led the process. All reviewers followed unified guidelines for identifying disinformation (see Appendix~\ref{app:guidelines} for details on experts review and guidelines). Label Studio \footnote{\href{https://github.com/HumanSignal/label-studio}{label-studio}}
 managed the annotation environment; disagreements were resolved via consensus or expert adjudication. { This phase achieved an overall Cohen's $\kappa = \mathbf{0.78}$, indicating strong inter-annotator agreement. To further validate reliability across multiple annotators, we computed Fleiss' $\kappa = \mathbf{0.76}$ on a stratified random sample of $n{=}500$ instances (proportional allocation across 13 categories, ensuring representation from both high-frequency and low-frequency sources). This multi-rater statistic confirms substantial agreement even when extending beyond pairwise comparisons.}

To gauge dependence on LLM assistance, we annotated an audit subset \emph{without} any LLM support: $n{=}200$ items stratified by category (13), outlet group (Top-10 vs.\ long-tail), and recency (last three months vs.\ earlier). Two experts independently labeled each item using the same guidelines, followed by adjudication. Agreement between audit labels and the main human-verified labels was $\kappa{=}\mathbf{0.82}$ (95\% CI [0.76, 0.88]) with \textbf{92.0\%} raw agreement ($n{=}200$).Example annotations appear in Figure~\ref{fig:disinfo-analysis}. Full prompt templates for dataset annotation and model evaluation are provided in Appendix~\ref{app:prompts}.

\begin{figure}[t]
    \centering
    \includegraphics[width=\textwidth]{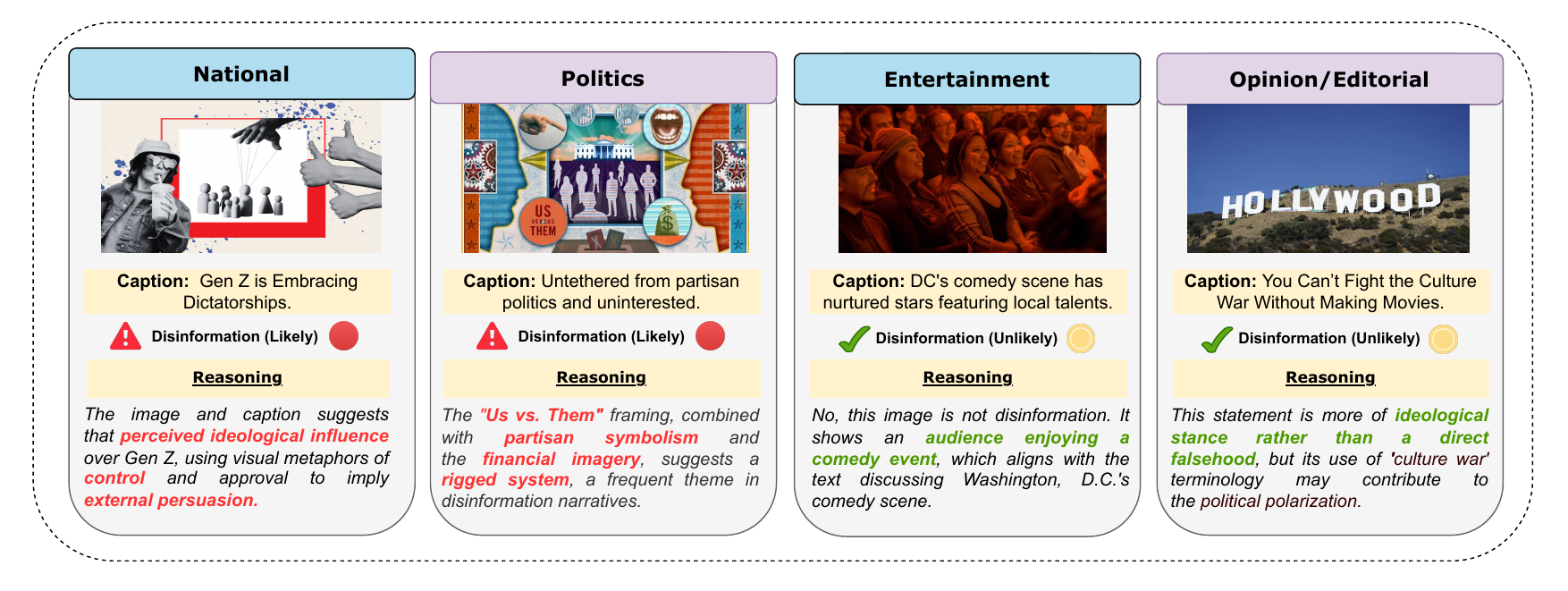}
    \caption{Disinformation trends across news categories based on GPT-4o narratives and confidence scores.}
    \label{fig:disinfo-analysis}
\end{figure}

\subsection{Benchmark Release}
We release the dataset under the \textbf{CC-BY-NC-SA 4.0} license \footnote{\href{https://creativecommons.org/licenses/by/4.0/deed.en}{CC BY 4.0 License}}
, permitting attribution-based, non-commercial use with share-alike terms. Personally identifiable information (PII) was scrubbed via regex and keyword filters; names of public figures were retained consistent with standard data governance norms. NSFW content was excluded via human review. We release the data and evaluation code to support reproducibility (Appendix~\ref{app:release}), including full prompt templates, model/version pins, decoding settings, and random seeds (see also Section~\ref{sec:experiment}).

\subsection{Exploratory Analysis}
\label{analysis}

\begin{table}[t]
    \centering
    \caption{Summary statistics of the \textsf{\textbf{\textsc{VLDBench}}} dataset.}
    \scriptsize
    \renewcommand{\arraystretch}{1.2} 
    \resizebox{\textwidth}{!}{
    \begin{tabular}{|p{6.2cm}|p{4.5cm}|}
        \hline
        \rowcolor{gray!15}\textbf{Metric} & \textbf{Value} \\ \hline
        Number of news sources & 58 \\
        Number of categories & 13 \\
        Modalities & 2 (text, image) \\ \hline
        Unique articles (paired text + image) & 31,339 \\
        Annotated instances (text-only) & 31,339 \\
        Annotated instances (multimodal: text + image) & 31,339 \\
        \textbf{Total annotated instances} & \textbf{62,678} \\ \hline
        Average article length & 1,168.7 words \\
        Image–caption coverage & 100\% (avg. 26.7 words) \\ \hline
        Disinformation (Likely / Unlikely) & 57.1\% / 42.9\% \\
        Collection period & May 2023 – September 2024 \\ \hline
    \end{tabular}
    }
    \label{tab:dataset_statistics}
\end{table}

\begin{figure}[h]
    \centering
    \includegraphics[width=0.7\textwidth]{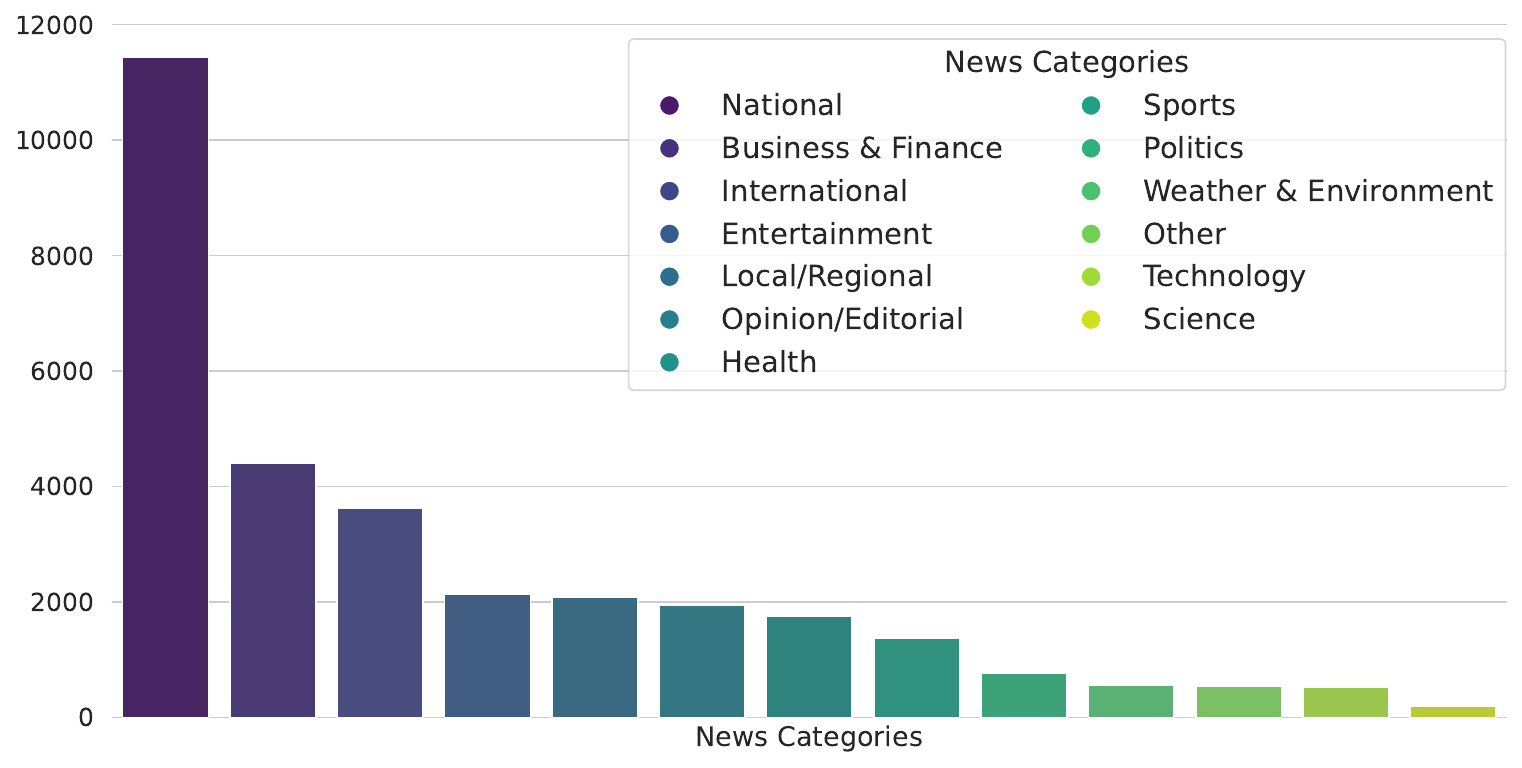}
    \caption{Category distribution with overlaps. Total unique articles = 31,339. Percentages exceed 100\% due to multi-category assignments. Political category omitted to reduce clutter.}
    \label{fig:news_categories}
\end{figure}

\noindent\textbf{Composition.}
\textsf{\textbf{\textsc{VLDBench}}} comprises 31{,}339 unique news articles (May~2023--Sep~2024) paired 1:1 with lead images (total 62{,}678 instances; 31{,}339 text-only and 31{,}339 text–image), spanning 13 categories (Figure~\ref{fig:news_categories}) and 58 outlets (Table~\ref{tab:sources}). Category assignments are multi-label in a minority of cases, so percentages in Figure~\ref{fig:news_categories} can exceed 100\% by design.

\noindent\textbf{Outlets.}
Coverage includes national and regional sources across North America and beyond, yielding a long-tail outlet distribution (Figure~\ref{fig:news_sources_distribution}); this diversity supports robustness to style and editorial variation.

\begin{figure}[h]
    \centering
    \includegraphics[width=0.7\textwidth]{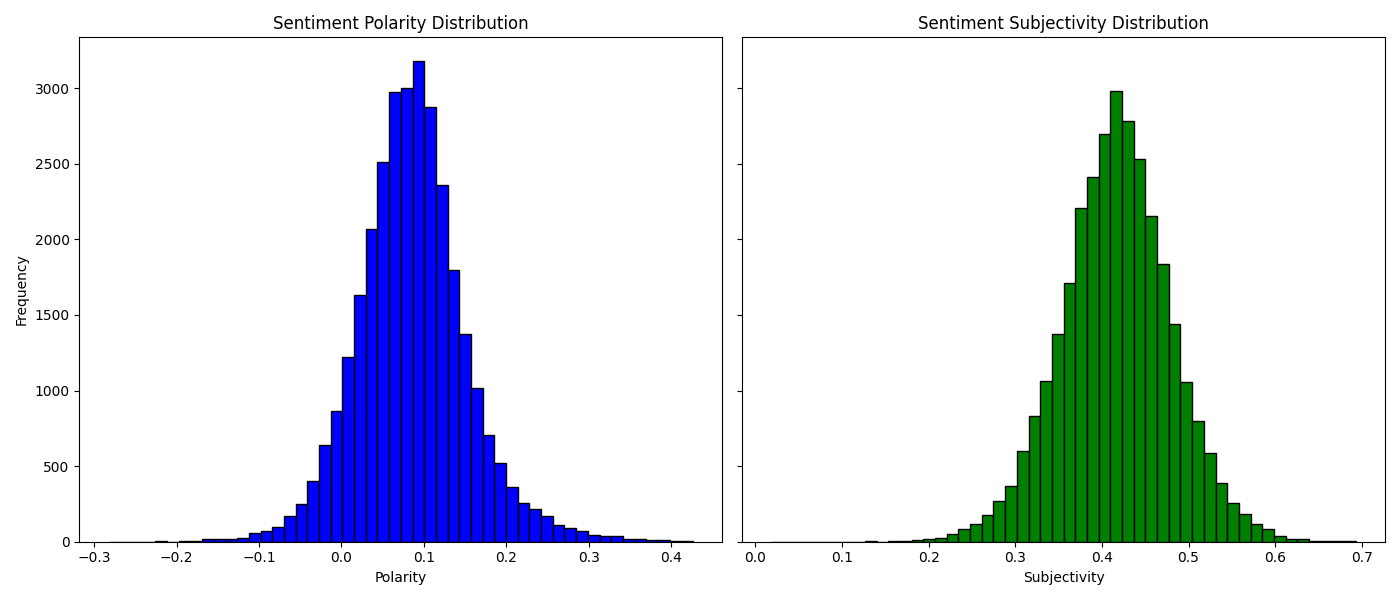}
    \caption{Sentiment polarity and subjectivity distributions. The dataset is mostly neutral or mildly positive, with a slight subjective tilt.}
    \label{fig:sentiment_distributions}
\end{figure}

\noindent\textbf{Sentiment \& Subjectivity.}
Aggregate polarity centers near neutral with a mild positive skew and slight subjective tilt (Figure~\ref{fig:sentiment_distributions}), which is expected for general-news corpora and helps avoid sentiment-only shortcuts.

\noindent\textbf{Class balance \& splits.}
Table~\ref{tab:dataset_statistics} reports label counts for \emph{Likely} vs.\ \emph{Unlikely Disinformation}. For prompting experiments, we use the full dataset to maximize coverage and robustness. For instruction-finetuning (IFT), we adopt a chronologically stratified 70/15/15 split by outlet, designed to mitigate leakage by ensuring that articles from the same outlet and closely adjacent publication dates are minimized across folds.

\begin{table}[h]
\centering
\caption{Grouped article sources by media outlet.}
\small
\renewcommand{\arraystretch}{1.2} 
\resizebox{\textwidth}{!}{
\begin{tabular}{|p{3.5cm}|p{11.5cm}|}
\hline
\rowcolor{gray!15}\textbf{Outlet} & \textbf{Sources / Subsections} \\ \hline
\textbf{AP News} & -- \\ \hline
\textbf{CBC} & CBC News; CBC Sports \\ \hline
\textbf{CBS} & Boston; Minnesota; New York; Miami; San Francisco; Colorado; Baltimore; Chicago; Pittsburgh; Sacramento; Los Angeles; Philadelphia \\ \hline
\textbf{Global News} & Toronto; Calgary; Edmonton; Halifax; British Columbia; Lethbridge; Guelph; Peterborough; Montréal; London; Kingston; Okanagan; Barrie; Ottawa; Winnipeg; Regina; Saskatoon; Hamilton \\ \hline
\textbf{Reuters} & UK; Canada; India; Reuters.com \\ \hline
\textbf{Washington Post} & Main; www-staging.washingtonpost.com \\ \hline
\textbf{The Guardian} & U.S.\ Edition \\ \hline
\textbf{USA Today} & WolverinesWire; Golfweek; Reviewed \\ \hline
\textbf{Fox News} & FOX News Radio \\ \hline
\textbf{CNN} & Underscored; International; Press Room \\ \hline
\textbf{The Economist} & Economist Impact \\ \hline
\end{tabular}
}
\label{tab:sources}
\end{table}

\section{Experimental Setup}
\label{sec:experiment}

\begin{table}[h]
\centering
\caption{Model families with language-only and vision-language variants.}
\scriptsize
\begin{tabular}{|l|p{5cm}|p{5cm}|}
\hline
\textbf{Family} & \textbf{Language-Only Models} & \textbf{Vision-Language Models} \\
\hline
Phi & Phi-3-mini-128k-Instruct & Phi-3-Vision-128k-Instruct \\
Mistral & Mistral-7B-Instruct-v0.3 & Pixtral \\
Qwen & Qwen2-7B-Instruct & Qwen2-VL-7B-Instruct \\
Intern & InternLM2-7B & InternVL2-8B \\
DeepSeek & DeepSeek-V2-Lite-Chat & DeepSeek-VL2-Small, Janus-Pro-7B \\
GLM & GLM-4-9B-Chat & GLM-4V-9B \\
Llama & Llama-3.1-8B, Llama-3.2-1B & Llama-3.2-11B-Vision \\
LLaVA & — & LLaVA-v1.5-Vicuna7B, LLaVA-v1.6-Mistral \\
Vicuna & Vicuna-7B-v1.5 & — (base used in LLaVA) \\
\hline
\end{tabular}

\label{tab:model_families}
\end{table}

\begin{table}[h]
\scriptsize
\centering
\caption{Hyperparameters used for large language models (LLMs) and vision–language models (VLMs).}
\label{tab:hyperparameters}
\renewcommand{\arraystretch}{1.2} 
\resizebox{\textwidth}{!}{%
\begin{tabular}{|p{0.28\linewidth}|p{0.62\linewidth}|}
\hline
\rowcolor{gray!15}\textbf{Hyperparameter} & \textbf{Value / Setting} \\ \hline
Temperature & 0.2 (more deterministic outputs) \\
Top-$p$ sampling & 0.9 (ablation range: 0.8–1.0) \\
Max generation length & 256 tokens (label + short rationale) \\ \hline
Effective batch size & 32 (via micro-batching; per-device size adapts to VRAM) \\
Context length & Up to 4{,}096 tokens (longest-first truncation) \\
Attention & FlashAttention-2 (when supported) \\
Checkpointing & Gradient checkpointing; save-best on dev F1; eval every 1{,}000 steps \\ \hline
Concurrency & Python \texttt{asyncio} dataloader with pinned memory and persistent workers \\ \hline
LLMs & \texttt{text\_content} \\
LVLMs& \texttt{text\_content + image} \\ \hline
Preprocessing & Resize longest side to 448\,px; center-crop if needed; normalize per model card \\
Formats & \texttt{.jpg}, \texttt{.jpeg}, \texttt{.png} (PNG optimized; JPEG quality 75) \\ \hline

\end{tabular}%
}
\end{table}

\begin{table}[h!]
\scriptsize
\centering
\caption{Perturbations for robustness evaluation. Each variant is evaluated independently; results are reported as $\Delta$F1 relative to the clean set.}
\label{tab:perturbation}
\renewcommand{\arraystretch}{1.2} 
\resizebox{\textwidth}{!}{%
\begin{tabular}{|p{0.1\linewidth}|p{0.15\linewidth}|p{0.65\linewidth}|}
\hline
\rowcolor{gray!15}\textbf{Axis} & \textbf{Type} & \textbf{Procedure} \\ \hline

\multirow{3}{*}{\textbf{Textual}} 
 & Synonym substitution & WordNet-based replacements via TextAttack \cite{morris2020textattack}; max 15\% token edits; preserve named entities. \\ \cline{2-3}
 & Misspelling noise & Random character-level edits on 20\% of words (Levenshtein distance $\leq$1); stopwords excluded. \\ \cline{2-3}
 & Negation flip & Insert ``not/never/no'' at clause-appropriate positions to invert polarity while preserving named entities. \\ \hline

\multirow{3}{*}{\textbf{Visual}} 
 & Blur & Gaussian blur with $\sigma \in \{0.5, 1.0, 1.5\}$; kernel auto-scaled to image size. \\ \cline{2-3}
 & Noise & Additive Gaussian noise ($\mu=0$, $\sigma \in \{0.05, 0.10\}$), clipped to valid pixel range. \\ \cline{2-3}
 & Resizing & Downsampled to 50\% and upsampled to 200\% (bicubic), preserving aspect ratio. \\ \hline

\multirow{2}{*}{\textbf{Cross-modal}} 
 & Mismatch & Swap image within same high-level category but from a different article; enforce semantic contradiction with caption/headline. \\ \cline{2-3}
 & Contradictory caption & Replace or append a short caption contradicting the image while leaving the article text unchanged. \\ \hline

\end{tabular}%
}
\end{table}

\paragraph{Hardware and Memory Optimization}
We used two GPU tiers: A100/H100 (80\,GB) for large VLMs and A10/3090/4090 (24–48\,GB) for ablations, balancing throughput and accessibility. To fit multimodal batches without accuracy loss, we used: (i) mixed precision (bf16/fp16 with automatic loss scaling) and FlashAttention-2 (where supported); (ii) gradient checkpointing and selective activation offloading; (iii) LoRA/QLoRA for instruction-tuning (IFT) on larger models; and (iv) paged KV-cache for long contexts. We tuned effective batch size via micro-batching and sequence packing, and capped input image resolution at 448\,px on the longer side for inference/IFT. Data I/O used cached, pre-resized images, pinned memory, persistent workers, and fused optimizers (e.g., fused AdamW when available). All runs pin model/version, tokenizer, prompts/decoding, and fixed seeds; we report mean$\pm$SD over three seeds.

\paragraph{Evaluated Models}
\emph{Language-only LLMs}: Phi-3-mini-128k-Instruct \cite{phi3mini2024}, Vicuna-7B-v1.5 \cite{zheng2023judging}, Mistral-7B-Instruct-v0.3 \cite{mistral20237b}, Qwen2-7B-Instruct \cite{yang2024qwen2}, Llama-3.1-8B-Instruct \cite{meta2024llama31}, Llama-3.2-1B-Instruct \cite{meta2024llama}, InternLM2-7B \cite{cai2024internlm2}, DeepSeek-V2-Lite-Chat \cite{liu2024deepseek}, and GLM-4-9B-Chat \cite{glm2024chatglm}. 
\emph{Vision-language models}: Phi-3-Vision-128k-Instruct \cite{phi3vision2024}, DeepSeek-Janus-Pro-7B \cite{chen2025janus}, DeepSeek-VL2-Small \cite{wu2024deepseek}, LLaVA-v1.5-Vicuna-7B \cite{NEURIPS2023LLaVA}, LLaVA-v1.6-Mistral \cite{NEURIPS2023LLaVA}, Pixtral \cite{mistral2024pix}, Qwen2-VL-7B-Instruct \cite{Qwen2VL}, Llama-3.2-11B-Vision \cite{meta2024llama}, GLM-4V-9B \cite{glm2024chatglm}, and InternVL2-8B \cite{chen2024internvl}.  Model familes are given in Table \ref{tab:model_families}.
To reduce bias from heterogeneous architectures, we (i) report within-family results (e.g., Llama, Qwen, Intern), and (ii) compare size-matched bins (1–8B, 8–15B, $>$15B). All runs share the same prompts, decoding settings, preprocessing, and seeds. We evaluate both zero-shot and instruction-tuned (IFT) settings. We configured models with the hyperparameters in Table~\ref{tab:hyperparameters}.

\paragraph{Evaluation Metrics}
We report accuracy, precision, recall, and F1 over the binary labels (\emph{Likely} vs.\ \emph{Unlikely Disinformation}). For \emph{reasoning quality}, we use GPT-4o as an automated judge following \cite{zheng2023judging}: the judge receives the item (with gold label hidden), the model’s label+rationale, and a concise rubric, then assigns 0–5 scores for relevance, correctness, and coherence. Items and systems are randomized to mitigate order effects; ties are broken by an independent re-evaluation. For robustness, we report the relative performance drop $\Delta$F1 across perturbed variants.

\paragraph{Robustness Setting}
To evaluate robustness, we design targeted perturbations across textual, visual, and cross-modal axes, simulating realistic disinformation tactics (manipulated captions, spelling noise, modality mismatches). Tables \ref{tab:perturbation}~ and Table \ref{tab:perturbations_desc} summarize the procedures.

\section{Benchmarking Language and Vision Models on VLDBench}
\label{sec:results}
In this section, we benchmark nine language-only LLMs and ten LVLMs on \textsf{\textbf{\textsc{VLDBench}}}. We report zero-shot and instruction-tuned (IFT) results and analyze robustness to textual/visual/cross-modal perturbations, scaling behavior, out-of-distribution generalization, and human-judged rationale quality. 
 All evaluations are performed on whole dataset ,but for IFT we use 70\% on train, 15\% on val, and reports on the 15\% test set. We focus on open-source LLMs and LVLMs to promote accessibility and transparency. GPT-4o was used only for dataset annotation and is deliberately excluded from evaluation to avoid same-model leakage and bias. We include both quantitative metrics and qualitative (human) judgments in zero-shot and IFT settings.
Our investigation centers on three questions:
(1)~Does multimodal (text+image) input improve detection over text alone?
(2)~Does instruction fine-tuning enhance generalization and robustness?
(3)~How vulnerable are models to adversarial perturbations in text and image modalities?

\begin{table}[t]
\centering
\caption{Zero-shot performance: language-only models (top) vs.\ large vision–language models (bottom) on \textbf{VLDBench}. \textbf{Bold} marks the best value in each column (higher is better). Means $\pm$ SD over three runs.}
\resizebox{0.95\textwidth}{!}{%
\begin{tabular}{lcccc}
\toprule
\textbf{Model} & \textbf{Accuracy (\%)} & \textbf{Precision (\%)} & \textbf{Recall (\%)} & \textbf{F1-Score (\%)}\\
\midrule
\multicolumn{5}{c}{\textit{Language-only LLMs}} \\
\midrule
Phi-3-mini-128k-Instruct \cite{phi3mini2024} & 57.15$\pm$0.29 & 55.12$\pm$0.31 & 58.21$\pm$0.35 & 56.62$\pm$0.28 \\
Vicuna-7B-v1.5 \cite{zheng2023judging} & 55.21$\pm$0.40 & 56.78$\pm$0.38 & 52.48$\pm$0.37 & 54.55$\pm$0.41 \\
Mistral-7B-Instruct-v0.3 \cite{mistral20237b} & 68.58$\pm$0.45 & 68.10$\pm$0.40 & 65.09$\pm$0.48 & 66.56$\pm$0.46 \\
Qwen2-7B-Instruct \cite{yang2024qwen2} & 69.92$\pm$0.50 & 68.34$\pm$0.44 & 69.10$\pm$0.49 & 68.72$\pm$0.53 \\
InternLM2-7B \cite{cai2024internlm2} & 51.19$\pm$0.32 & 49.85$\pm$0.41 & 51.16$\pm$0.42 & 50.50$\pm$0.36 \\
DeepSeek-V2-Lite-Chat \cite{liu2024deepseek} & 51.96$\pm$0.49 & 52.61$\pm$0.39 & 53.53$\pm$0.40 & 51.96$\pm$0.44 \\
GLM-4-9B-Chat \cite{glm2024chatglm} & 51.14$\pm$0.55 & 60.19$\pm$0.51 & 53.28$\pm$0.47 & 56.52$\pm$0.50 \\
LLaMA-3.1-8B-Instruct \cite{meta2024llama31} & 68.21$\pm$0.42 & 62.13$\pm$0.43 & 62.10$\pm$0.45 & 62.11$\pm$0.40 \\
LLaMA-3.2-1B-Instruct \cite{meta2024llama} & 70.29$\pm$0.36 & 69.78$\pm$0.33 & 68.92$\pm$0.37 & 69.35$\pm$0.39 \\
\midrule
\multicolumn{5}{c}{\textit{Large Vision–Language Models (LVLMs)}} \\
\midrule
Phi-3-Vision-128k-Instruct \cite{phi3vision2024} & 64.18$\pm$0.43 & 63.18$\pm$0.40 & 62.88$\pm$0.44 & 63.03$\pm$0.38 \\
LLaVA-v1.5-Vicuna-7B \cite{NEURIPS2023LLaVA} & 72.32$\pm$0.48 & 68.12$\pm$0.49 & 64.88$\pm$0.41 & 66.46$\pm$0.47 \\
LLaVA-v1.6-Mistral-7B \cite{NEURIPS2023LLaVA} & 72.38$\pm$0.42 & 70.18$\pm$0.46 & 70.03$\pm$0.39 & 70.10$\pm$0.42 \\
Pixtral \cite{mistral2024pix} & 70.18$\pm$0.39 & 72.32$\pm$0.37 & 70.23$\pm$0.34 & 71.26$\pm$0.38 \\
Qwen2-VL-7B-Instruct \cite{Qwen2VL} & 67.28$\pm$0.33 & 69.18$\pm$0.42 & 68.45$\pm$0.35 & 68.81$\pm$0.32 \\
InternVL2-8B \cite{chen2024internvl} & 63.57$\pm$0.36 & 68.34$\pm$0.38 & 65.10$\pm$0.34 & 66.88$\pm$0.40 \\
DeepSeek-VL2-Small \cite{wu2024deepseek} & 68.15$\pm$0.31 & 64.84$\pm$0.37 & \textbf{75.67}$\pm$0.45 & 69.84$\pm$0.34 \\
DeepSeek Janus-Pro-7B \cite{chen2025janus} & 70.04$\pm$0.42 & 69.97$\pm$0.43 & 74.14$\pm$0.46 & 71.99$\pm$0.41 \\
GLM-4V-9B \cite{glm2024chatglm} & 62.13$\pm$0.44 & 63.35$\pm$0.42 & 68.82$\pm$0.47 & 66.82$\pm$0.41 \\
LLaMA-3.2-11B-Vision \cite{meta2024llama} & \textbf{74.82}$\pm$0.39 & \textbf{72.62}$\pm$0.45 & 72.28$\pm$0.36 & \textbf{72.45}$\pm$0.42 \\
\bottomrule
\end{tabular}%
}

\label{tab:main}
\end{table}

\subsection{Multimodal Models Surpass Unimodal Baselines}
\label{sec:multimodal_vs_unimodal}
We compare the zero-shot performance of language-only LLMs and vision–language LVLMs on our disinformation benchmark. As shown in Table~\ref{tab:main}, LVLMs consistently outperform their unimodal counterparts. For instance, LLaMA-3.2-11B-Vision surpasses its text-only family member (LLaMA-3.2-1B) in both accuracy and F1. Models such as Phi-3-Vision, LLaVA-v1.5/v1.6, Pixtral, InternVL2, DeepSeek-VL2, and GLM-4V also show notable gains over their respective LLM baselines. Notably, LLaMA-3.2-11B-Vision attains the best overall accuracy (74.82\%) and F1 (72.45\%).

Performance improvements are particularly pronounced in some families. LLaVA-v1.5-Vicuna-7B improves accuracy over its base Vicuna-7B by +17.1 percentage points (\textasciitilde\textbf{31}\% relative), underscoring the role of visual grounding. One exception is Qwen2-VL-7B, which underperforms slightly relative to Qwen2-7B (67.28\% vs.\ 69.92\% accuracy), suggesting modality integration remains architecture- and alignment-dependent. While strong LLMs (e.g., LLaMA-3.2-1B at 70.29\% accuracy) remain competitive, LVLMs generally achieve higher recall, peaking at 75.67\% for DeepSeek-VL2-Small, which is critical for minimizing missed detections in adversarial settings. Standard deviations remain $<1\%$ across metrics, indicating stable performance.

\subsection{Instruction Fine-Tuning (IFT) on \textsf{\textbf{\textsc{VLDBench}}} Improves Performance}
\label{sec:instruction_ft}
\begin{figure}[h!]
    \centering
    \includegraphics[width=0.8\textwidth]{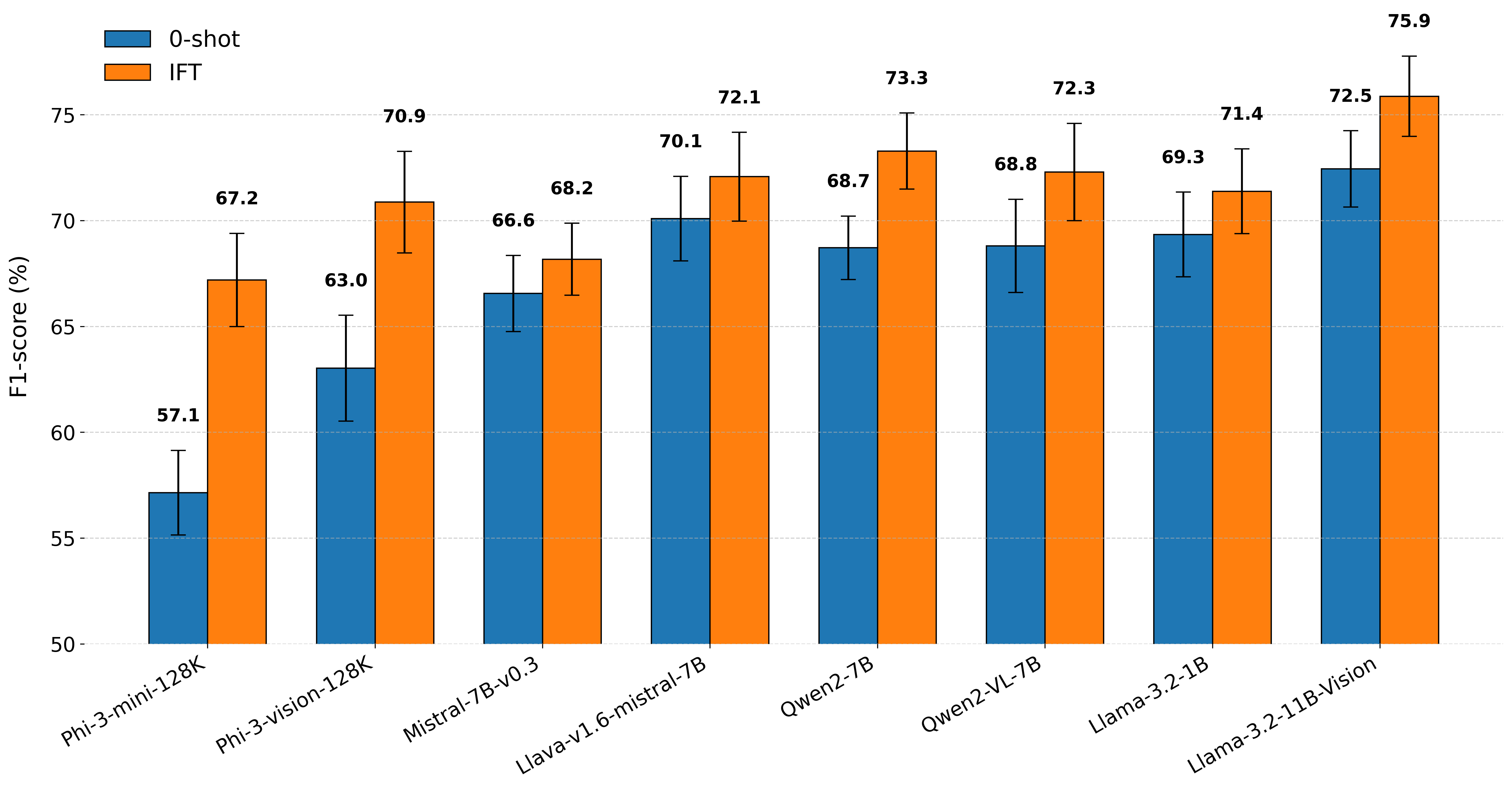}
    \caption{Zero-shot vs.\ instruction-tuned (IFT) performance with 95\% confidence intervals over 3 independent runs.}
    \label{fig:ift}
\end{figure}

We apply IFT to representative models using the training subset of \textsf{\textbf{\textsc{VLDBench}}}. As shown in Figure~\ref{fig:ift}, IFT consistently improves performance over zero-shot baselines. For example, Phi-3-mini-128k-Instruct improves by +10.1 percentage points in F1 (from 57.1\% to 67.2\%), while its vision counterpart Phi-3-Vision-128k-Instruct rises from 63.0\% to 70.9\%. Notably, LLaMA-3.2-11B-Vision attains the highest post-IFT F1 at 75.9\%, reinforcing the value of task-specific multimodal alignment.

These gains indicate that fine-tuning helps models capture disinformation-specific cues from multimodal content. The improvements are consistent across architectures suggests that IFT is a robust strategy for both unimodal and multimodal detectors on this benchmark.

\subsection{Adversarial Robustness: Combined Modality is More Vulnerable}
\label{sec:resilience}
We evaluate robustness with four controlled perturbation types:
(1) \textit{Text perturbations (T-P)}: synonym substitutions, typos, and negations; 
(2) \textit{Image perturbations (I-P)}: blur, noise, and resizing; 
(3) \textit{Cross-modal misalignment (C-M)}: mismatched or contradictory text–image pairs; and 
(4) \textit{Both-modality perturbations (B-P)}: simultaneous text and image changes.
Unless noted, robustness results use the zero-shot setting; see Appendix~\ref{app:perturbations} for full details. Our results are discussed next as:

\begin{table}[h]
\centering
\scriptsize
\caption{
Model robustness under controlled perturbations. Each value (e.g., T-P = $-3.2$) indicates the absolute F1 score drop (in percentage points) compared to the model’s original performance. Perturbation types include: T-P (text perturbation), I-P (image perturbation), C-M (cross-modal misalignment), and B-P (both modalities perturbed). Larger drops indicate greater performance degradation. \textbf{Bold} values highlight the weakest performance per category. Red and orange gradients visually encode severity (darker = greater drop). Overall, LLMs exhibit greater vulnerability to text perturbations (T-P), while LVLMs are more affected by cross-modal and combined attacks.
}

\renewcommand{\arraystretch}{1}
\setlength{\tabcolsep}{4pt}
\resizebox{0.78\textwidth}{!}{
\begin{tabular}{|l|c|c|c|c|c|c|}
\hline
\rowcolor{gray!15}
\textbf{Model} & \textbf{Orig. F1} & \textbf{T-P} & \textbf{I-P} & \textbf{C-M} & \textbf{B-P} & \textbf{Avg. Drop} \\
\hline
\multicolumn{7}{|c|}{\textbf{Language-Only LLMs}} \\
\hline
LLaMA-3.2-1B & 69.35 & $-3.2$ & $-2.6$ & $-6.3$ & \cellcolor{red!35} $-11.2$ & \cellcolor{red!25} 5.83 \\
Mistral-7B & 66.56 & $-3.5$ & $-2.5$ & $-6.2$ & \cellcolor{red!34} $-11.0$ & \cellcolor{red!24} 5.80 \\
InternLM2-7B & 66.68 & $-2.7$ & $-3.1$ & $-6.8$ & \cellcolor{red!40} $-11.8$ & \cellcolor{red!30} 6.10 \\
Vicuna-7B & 54.55 & $-4.1$ & $-2.0$ & $-6.0$ & \cellcolor{red!30} $-10.3$ & \cellcolor{red!22} 5.60 \\
Qwen2-7B & 68.72 & $-3.0$ & $-2.2$ & $-5.9$ & \cellcolor{red!28} $-10.1$ & \cellcolor{red!20} 5.30 \\
Phi-3-mini & 55.71 & $-3.2$ & $-1.8$ & $-5.5$ & \cellcolor{orange!25} $-9.4$ & \cellcolor{orange!18} 4.98 \\
DeepSeek-V2-Lite & 51.96 & $-2.7$ & $-3.2$ & $-6.3$ & \cellcolor{red!28} $-10.1$ & \cellcolor{red!20} 5.58 \\
\hline
\multicolumn{7}{|c|}{\textbf{Large Vision-Language Models (LVLMs)}} \\
\hline
LLaMA-3.2-11B-Vision & 72.45 & $-2.0$ & $-2.8$ & $-5.6$ & \cellcolor{red!28} $-10.2$ & \cellcolor{red!20} 5.15 \\
LLaVA-v1.6-Mistral7B & 70.10 & $-2.1$ & $-2.9$ & $-5.8$ & \cellcolor{red!32} $-10.7$ & \cellcolor{red!24} 5.38 \\
InternVL2-8B & 66.68 & $-2.7$ & $-3.1$ & $-6.8$ & \cellcolor{red!40} $-11.8$ & \cellcolor{red!30} 6.10 \\
LLaVA-v1.5-Vicuna7B & 70.10 & $-2.8$ & $-3.4$ & $-7.0$ & \cellcolor{red!40} $-12.6$ & \cellcolor{red!30} 6.45 \\
Qwen2-VL-7B & 65.86 & $-2.3$ & $-3.0$ & $-6.4$ & \cellcolor{red!35} $-11.5$ & \cellcolor{red!25} 5.80 \\
Phi-3-Vision & 63.03 & $-2.5$ & $-3.2$ & $-6.1$ & \cellcolor{red!35} $-11.2$ & \cellcolor{red!25} 5.75 \\
Pixtral & 71.26 & $-1.9$ & $-2.7$ & $-5.4$ & \cellcolor{orange!30} $-9.8$ & \cellcolor{orange!20} 4.95 \\
DeepSeek-VL2-small & 69.84 & $-2.2$ & $-3.0$ & $-6.5$ & \cellcolor{red!32} $-10.7$ & \cellcolor{red!24} 5.60 \\
\hline
\end{tabular}
}

\label{tab:perturbations1}
\end{table}

\begin{table}[h!]
\centering
\caption{Adversarial accuracy and absolute drop ($\Delta$); \textsuperscript{§} indicates combined text–image attacks.}

\footnotesize
\resizebox{0.78\textwidth}{!}{
\begin{tabular}{@{}llr@{}}
\toprule
\textbf{Model/Condition} & \textbf{Scenario} & \textbf{Accuracy\ ($\Delta$\%)}\\ 
\midrule
\textbf{LLaMA-3.2-1B} (Original) & No adversaries & \cellcolor{green!30} 75.90 (—) \\  
\textbf{LLaMA-3.2-1B} (Text) & Synonym/misspelling/negation & \cellcolor{yellow!40} 60.85 ($\Delta$15.05) \\  
\textbf{LLaMA-3.2-11B} (Combined) & Text+image attacks\textsuperscript{§} & \cellcolor{red!40} 53.54 ($\Delta$22.36) \\  
\bottomrule
\end{tabular}
}
\label{tab:adversarial_performance}
\end{table}

\begin{figure}[h!]
  \centering
  \begin{subfigure}[t]{0.45\textwidth}
    \centering
    \includegraphics[width=1.05\linewidth]{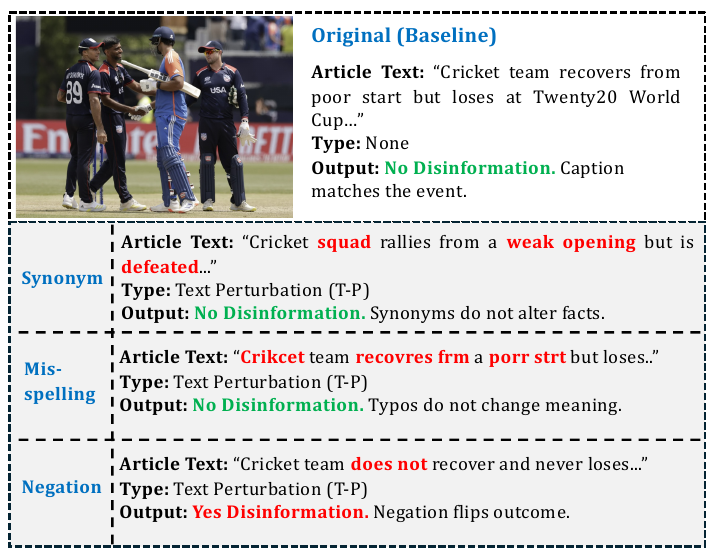}
    \caption{\textbf{Textual perturbations.} using test synonyms substitutions, misspellings, and negation; negation induces the sharpest drop.}
    \label{fig:perturbation_text}
    \vspace{-1em}
  \end{subfigure}\hfill
  \begin{subfigure}[t]{0.45\textwidth}
    \centering
    \includegraphics[width=0.8\linewidth]{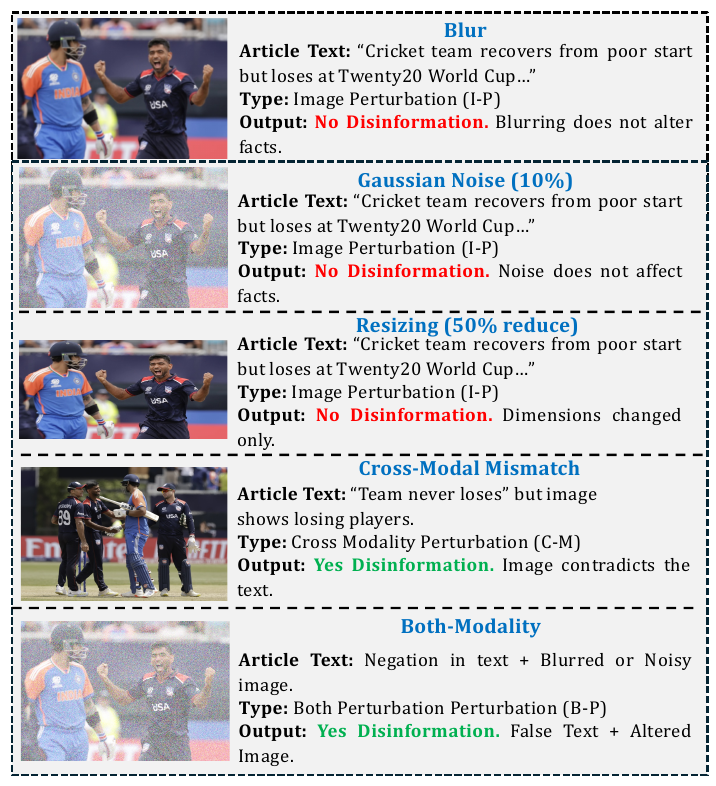}
    \caption{\textbf{Visual perturbations} using blur, additive noise, and resizing. Visual corruption alone is less damaging than cross-modal mismatches.}
     \vspace{-1em}
    \label{fig:perturbation_image}
  \end{subfigure}
  \caption{Illustration of (a) text-only and (b) image-only perturbation strategies. Cross-modal and both-modality variants are described in Appendix~\ref{app:perturbations}.}
  \vspace{-1em}
  \label{fig:perturbations_both}
\end{figure}

\textbf{Single-modality attacks.}
We first test robustness under isolated text or image corruption (zero-shot). 
Textual attacks include synonym substitutions, misspellings (20\% of words; bounded edit distance), and negation insertions; 
visual attacks include Gaussian blur (3$\times$3), additive noise ($\mu{=}0$, $\sigma{=}0.1$), and resizing (50\% / 200\%). 
Table~\ref{tab:perturbations1} shows that LVLMs generally degrade less than LLMs under single-modality attacks, which indicates that complementary signals from images help in disinformation detection when the text is noisy. 

\textbf{Cross-modal and combined attacks.}
We observe much performance drops under cross-modal mismatches and both-modality perturbations. For example, Phi-3-Vision-128k-Instruct declines by 11.2 points under combined attacks, and LLaMA-3.2-11B-Vision by 10.2 points (Table~\ref{tab:perturbations1}). This result reveals a big vulnerability when both channels are distorted. 

{ 
Using the model-wise \emph{absolute} macro-F1 drops (pp) in Table~\ref{tab:perturbations1}, the cross-modal misalignment drop (C-M) is strongly correlated with the combined-attack drop (B-P) across all models (Pearson $r{=}0.88$; Appendix Fig.~\ref{fig:cm-bp-corr}), linking cross-modal mismatch to the observed robustness degradation. 
Within LVLMs only ($n{=}8$), text-perturbation sensitivity (T-P; largely driven by negation) also tracks combined-attack vulnerability (Pearson $r{=}0.95$).
}

\textbf{Instruction-tuned models under combined attacks.}
The trend persists for IFT models: as summarized in Table~\ref{tab:adversarial_performance}, LLaMA-3.2-11B-vision exhibits a substantial \textbf{22.36} point accuracy drop under combined attacks (text+image). 
Despite higher clean accuracy, tuned LVLMs remain sensitive to coordinated multimodal perturbations, suggesting that future work should pair IFT with robustness-aware training and cross-modal consistency regularization.

\subsection{Scalability Improves Model Performance}
\label{sec:scalability}
\begin{table}[h]
\centering
\caption{Performance metrics of vision models scaled up with increased weights and parameters, indicating performance improvements.}
\resizebox{0.95\textwidth}{!}{
\begin{tabular}{|l|c|c|c|c|c|}
\hline
\textbf{Model} & \textbf{Accuracy (\%)} & \textbf{Precision (\%)} & \textbf{Recall (\%)} & \textbf{F1-Score (\%)} & \textbf{Performance Increase (\%)} \\ \hline
Llama-3.2-11B-Vision & 74.82 & 72.62 & 72.28 & 72.45 & - \\ \hline
Llama-3.2-90B-Vision & 76.8  & 78.43 & 78.06 & 78.24 & 3.42\% (from 11B) \\ \hline
InternVL2-8B & 63.57 & 68.34 & 65.1  & 66.68 & - \\ \hline
InternVL2-26B & 68.62 & 73.8  & 70.31 & 72.01 & 8.05\% (from 8B) \\ \hline
\end{tabular}
}

\label{tab:scale}
\end{table}

We evaluate the impact of scaling on LVLMs' performance by comparing smaller and larger variants within the same model family. In this experiment, we focused on representative models due to computational constraints. As shown in Table~\ref{tab:scale}, increasing model size consistently improves performance across accuracy, precision, recall, and F1 score.
For instance, LLaMA-3.2-Vision exhibits a +3.42\% gain in F1 when scaled from 11B to 90B parameters, while InternVL2 achieves a +8.05\% improvement when scaling from 8B to 26B. These consistent gains indicate that larger models are better equipped to capture fine-grained multimodal cues. This trend suggests that scaling enhances generalization and robustness, particularly in complex disinformation detection scenarios that require joint reasoning over visual and textual inputs.


\subsection{Domain Shift Impacts Performance though Multimodal Models Remain Stronger}
We further assess multimodal robustness on the r/Fakeddit benchmark, a Reddit-sourced dataset of text--image pairs with distantly supervised labels (2-, 3-, and 6-way) \cite{nakamura-etal-2020-fakeddit}.
{We evaluate cross-domain robustness on r/Fakeddit in the binary fake/true setting. Importantly, these labels are not intent-equivalent to our disinformation definition: in this context, ``fake'' often reflects satire, fiction, or other non-deceptive content, while ``true'' can still be used in misleading ways via selective framing or missing context.
For example, a satirical meme may be labeled ``fake'' despite lacking deceptive intent, whereas a factually accurate post may be labeled ``true'' while still misleading readers through a cropped image or omission of crucial context.
We therefore report r/Fakeddit results as an out-of-distribution stress test rather than an intent-aligned disinformation evaluation.}

For evaluation, we sample $N=1000$ class-balanced items \emph{per seed} (500 per class; discarding entries with missing images) using three fixed random seeds (0, 1, 2), obtain images via the official Fakeddit script, and resize each image to $\leq 512$\,px on the longer side, yielding $3000$ total evaluations across seeds.
{We report mean$\pm$SD over the three runs and evaluate (i) our best VLM on image+text and (ii) our strongest text model on text only. All runs are zero-shot and reuse the same prompts/decoding as our main experiments.
}

\begin{table}[h]
\centering
\caption{In-domain (VLDBench) vs.\ external out-of-distribution (OOD) (r/Fakeddit, binary, zero-shot). 
Accuracy and macro-F1 are mean~$\pm$~SD over three seeds. 
$\Delta$ F1 is the change in macro-F1 relative to the same model on VLDBench.}
\label{tab:ood}
\setlength{\tabcolsep}{6pt}\renewcommand{\arraystretch}{1.12}
\resizebox{\textwidth}{!}{
\begin{tabular}{l l l r r r}
\toprule
Dataset & Model & Modality & Acc (\%) & Macro-F1 (\%) & $\Delta$F1 \\
\midrule
VLDBench (in-domain) & LLaMA-3.2-11B-Vision  & image+text & 74.82 $\pm$ 0.39 & 72.45 $\pm$ 0.42 & 0.00 \\
VLDBench (in-domain) & LLaMA-3.2-1B-Instruct & text only  & 70.29 $\pm$ 0.36 & 69.35 $\pm$ 0.39 & 0.00 \\
\midrule
r/Fakeddit (OOD)     & LLaMA-3.2-11B-Vision  & image+text & 71.40 $\pm$ 0.60 & 66.80 $\pm$ 0.70 & \cellcolor{red!15}$-5.65$ \\
r/Fakeddit (OOD)     & LLaMA-3.2-1B-Instruct & text only  & 66.20 $\pm$ 0.70 & 62.30 $\pm$ 0.80 & \cellcolor{red!25}$-7.05$ \\
\bottomrule
\end{tabular}
}
\vspace{2pt}
\footnotesize {Fakeddit “fake/true” serves as a stress test rather than an intent-equivalent label.}

\end{table}

The results in Table~\ref{tab:ood} show that, on r/Fakeddit (binary, zero-shot), both models degrade relative to VLDBench. 
Macro-F1 falls from $72.45\pm0.42$ to $66.80\pm0.70$ for the LVLM (LLaMA-3.2-11B-Vision) and from $69.35\pm0.39$ to $62.30\pm0.80$ for the text model (LLaMA-3.2-1B-Instruct); accuracy declines are comparable. 
Despite this shift, the LVLM remains stronger on r/Fakeddit ($66.8$ vs.\ $62.3$ macro-F1), indicating that image information is still useful on social-media–style data. 
The reductions exceed run-to-run variability (SD $\leq 0.8$), consistent with a real domain-shift effect. 
Given the differences between newsroom articles and Reddit posts (shorter, noisier text; meme-like imagery; distantly supervised labels), we treat r/Fakeddit as an external stress test that complements our in-domain outlet and temporal hold-outs.

\subsection{Human Evaluation Demonstrates Reliability and Reasoning Depth}  
\label{sec:human_eval}  

To evaluate how well multimodal models justify their disinformation predictions, we conducted a human study focusing on two dimensions: predictive reliability and explanation quality. Specifically, we assessed two instruction-tuned LVLMs: \textit{LLaMA-3.2-11B-V} and \textit{LLaVA-v1.6}, on a balanced test set of 500 samples (250 disinformation, 250 neutral). Each model output included a binary prediction (likely/unlikely disinformation) and a corresponding rationale.

Three independent annotators, blinded to model identity, scored each output on two qualitative dimensions:  
\textbf{Prediction Correctness (PC)}: how well the model’s prediction aligned with the ground truth, and  
\textbf{Reasoning Clarity (RC)}: how coherent and interpretable the model’s explanation was.   Both were rated on a 1–5 Likert scale.

\begin{table}[h]
    \centering
    \small
    \resizebox{1\columnwidth}{!}{
    \begin{tabular}{lccc}
    \toprule
    \textbf{Model} & \textbf{Accuracy (\%)} & \textbf{Prediction Correctness (PC)} & \textbf{Reasoning Clarity (RC)} \\
    \midrule
    LLaMA-3.2-11B-V & 75.2 & 3.8 $\pm$ 0.7 & 3.5 $\pm$ 0.8 \\
    LLaVA-v1.6 & 72.0 & 3.1 $\pm$ 0.9 & 3.0 $\pm$ 1.0 \\
    \bottomrule
    \end{tabular}}
    \caption{Human evaluation on a 500-sample test set. PC = prediction correctness, RC = reasoning clarity. Scores are on a 1–5 scale. Mean $\pm$ standard deviation reported.}
    \label{tab:reasons}
\end{table}

\begin{figure}[t] 
    \centering
    \includegraphics[width=\columnwidth]{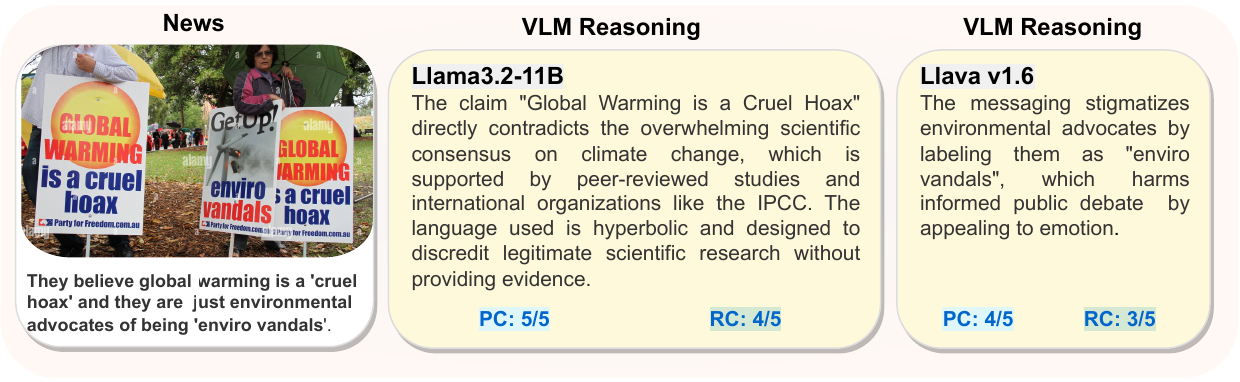} 
   \caption{Human evaluation results. Annotators assessed both prediction correctness and explanation clarity. Example shows difference in rationale quality between models.}
    \label{fig:reasoning-example}
\end{figure}

As shown in Table~\ref{tab:reasons}, LLaMA-3.2-11B-V achieved the highest accuracy and received higher ratings for both correctness and clarity. In contrast, LLaVA-v1.6, though reasonably accurate, often produced rationales lacking detail or coherence. These results suggest that larger, more recent LVLMs not only perform better but also offer more interpretable justifications for disinformation detection. A qualitative example is shown in Figure \ref{fig:reasoning-example}.

\subsection{Error Analysis}

We examine where models fail to complement the robustness results. 
Table~\ref{tab:slices} reports macro-F1 by news category, outlet group, and recency. Performance is lower for categories with shorter, more colloquial text and meme-style imagery (e.g., Entertainment), and for long-tail outlets with atypical style. 
We also observe recurrent failure modes: (1) out-of-date or misattributed visuals (an older image framed as new); (2) caption-image contradiction (text asserts $X$ while the image suggests $\neg X$); and (3) implication (true facts arranged to imply a false causal link). These patterns are consistent with the larger drops under cross-modal and both-modality perturbations (Table \ref{tab:perturbations1}), indicating that misalignment and visual reuse are primary drivers of error.

\begin{table}[h]
\small
\centering
\caption{Slice performance (macro-F1, \%). Means~$\pm$~SD over three seeds for the strongest text-only model and VLM on the same test split. Slices are label-preserving subsets: Politics/Entertainment; Top-10 vs.\ long-tail outlets (by volume); Recent = last three calendar months in the test set.}
\setlength{\tabcolsep}{5pt}\renewcommand{\arraystretch}{1}
\begin{tabular}{lrr}
\toprule
Slice & Text-only & Image+Text \\
\midrule
Category: Politics            & 68.2$\pm$0.6 & \textbf{71.5}$\pm$0.5 \\
Category: Entertainment       & 63.1$\pm$0.7 & \textbf{66.0}$\pm$0.6 \\
Outlet: Top-10                & 70.4$\pm$0.5 & \textbf{73.0}$\pm$0.4 \\
Outlet: Long-tail             & 64.9$\pm$0.8 & \textbf{67.2}$\pm$0.7 \\
Month: Recent (last 3 months) & 68.7$\pm$0.6 & \textbf{71.1}$\pm$0.6 \\
\bottomrule
\end{tabular}
\label{tab:slices}
\end{table}

As shown in Table \ref{tab:slices}, the LVLM exceeds the text-only model by roughly 2–3 macro-F1 points. Entertainment lags behind Politics, and long-tail outlets underperform Top-10 outlets, indicating sensitivity to source variability. 
Overall, the slice results and the robustness suite point to cross-source/style factors and cross-modal misalignment as the main sources of error.

Next, we discuss the AI risk mitigation and goverance alignment in \textsf{\textbf{\textsc{VLDBench}}}.
 \subsection{AI Risk Mitigation and Governance Alignment in \textsf{\textbf{\textsc{VLDBench}}}}
\label{sec:arr}
\textsf{\textbf{\textsc{VLDBench}}} has been developed with reference to the AI risks mentioned in MIT AI Risk Repository \cite{slattery_2024ai}, thereby situating its evaluation design within a recognized governance framework. This alignment ensures that the benchmark not only provides standard performance metrics but also systematically addresses risks of societal concern such as privacy, disinformation, fairness, robustness, and transparency. In this way, \textsf{\textbf{\textsc{VLDBench}}} operates at the intersection of technical benchmarking and risk-informed governance, a connection that has often remained underexplored in existing datasets.

\begin{figure}[h]
    \centering
    \includegraphics[width=0.7\textwidth]{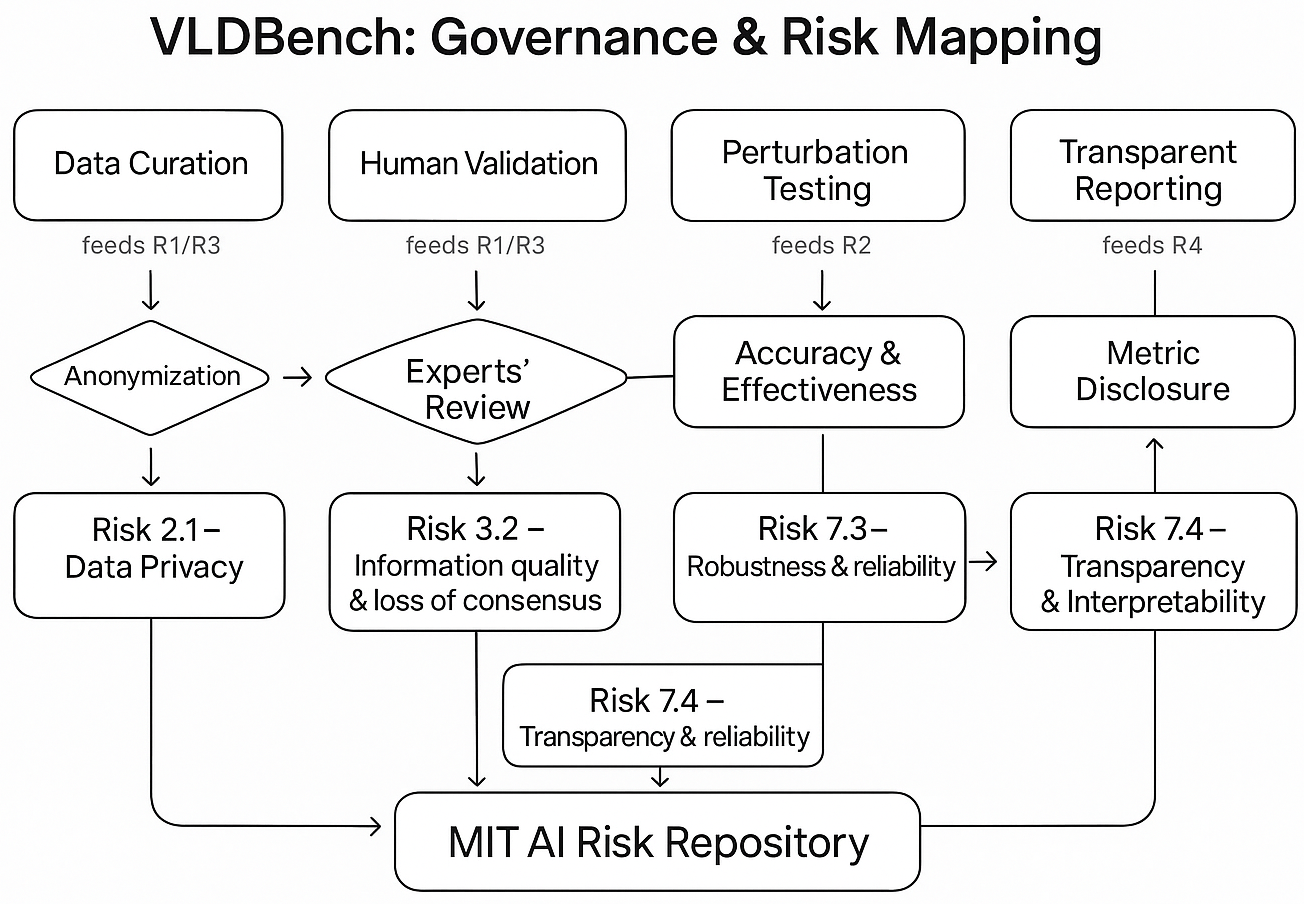}
   \caption{Governance and risk mapping in \textsf{\textbf{\textsc{VLDBench}}}, linking pipeline processes (data curation, validation, perturbation testing, and reporting) to risks in the MIT AI Risk Repository. Arrows illustrate how governance steps (e.g., anonymization, expert review, accuracy checks, metric disclosure) trace directly to identified risks: data privacy (2.1), information quality and consensus (3.2), discrimination and misrepresentation (1.1/1.2), robustness and reliability (7.3), and transparency and interpretability (7.4). All risks converge into the MIT AI Risk Repository, ensuring regulatory alignment and traceability.}

    \label{fig:vldbench_ai_risk_map}
\end{figure}

Figure~\ref{fig:vldbench_ai_risk_map} illustrates the mapping between pipeline components and the corresponding governance risk areas. Data curation and anonymization mitigate privacy risks; validation and expert review target information quality and the prevention of misrepresentation; perturbation testing directly addresses robustness and reliability; and disclosure of rationales and metrics enhances transparency and interpretability.

To operationalize this alignment, we introduce a \emph{Risk Scorecard} that complements accuracy and macro-F1 with four lightweight indicators derived from the MIT repository. These indicators are computed on the same predictions and test splits as the main experiments, requiring no retraining or model modification. Their purpose is not to propose new evaluation methods but rather to instantiate governance-relevant dimensions in measurable form. Specifically:

\noindent\textbf{(R1) Disinformation risk.} Share of \emph{Likely Disinformation} items missed (false negatives), reported overall and across three sensitive domains: Elections/Politics, Health/Science, and Public-safety/Crisis. Higher values indicate greater exposure to harmful content.\\
\textbf{(R2) Robustness risk.} Performance degradation under stress conditions. We compute the drop in macro-F1 under cross-modal mismatches and both-modality perturbations, averaging across stress types. Larger drops reflect greater brittleness.\\
\textbf{(R3) Disparity risk.} Variability in performance across content domains. Measured as the gap between the highest and lowest macro-F1 among the slices above, with larger values suggesting uneven reliability.\\
\textbf{(R4) Transparency.} Human-assessed clarity of model rationales, scored as the proportion judged “clear and evidence-based” according to the rubric introduced in Sec.~3.5. Higher values indicate stronger interpretability.

{
For R1--R3, we report each indicator as mean$\pm$SD over three fixed random seeds (0, 1, 2) using the same evaluation split, prompts, and decoding.
For R4 (Transparency), we binarize rationale quality using the human Reasoning Clarity (RC) ratings in Sec.~5.6: a rationale \emph{passes} if its average RC score is $\ge 4$ (out of 5), and we report the pass rate with a 95\% bootstrap CI over the 500-sample human-evaluation set.}


\section{Discussion}

\subsection{Impact}
\label{sec:social-impact }

Disinformation threatens democratic institutions, public trust, and social cohesion, and generative AI has amplified these risks by enabling sophisticated \emph{multimodal} campaigns. These campaigns exploit cultural, political, and linguistic nuances, demanding solutions that extend beyond purely technical approaches. \textsf{\textbf{\textsc{VLDBench}}} addresses this challenge by establishing a multimodal benchmark for disinformation detection that combines text and image analysis with explicit ethical safeguards. To ensure cultural sensitivity, annotations account for regional and contextual variation; bias audits and human–AI hybrid validation mitigate stereotyping risks. Ground-truth labels are derived from fact-checked sources with provenance tracking, aligning the benchmark with transparency and accountability principles.

The societal value of \textsf{\textbf{\textsc{VLDBench}}} lies in its dual role as a technical resource and a catalyst for interdisciplinary collaboration. By open-sourcing the benchmark and models, we lower barriers for researchers and practitioners in resource-constrained settings, democratizing access to state-of-the-art detection tools. Quantitative performance gaps, such as vulnerability to adversarial attacks—also surface systemic risks in current systems, incentivizing safer and more reliable alternatives. The benchmark is designed to foster partnerships across academia, industry, journalism, and policy, bridging the gap between computational research and real-world impact.

\subsection{Limitations and Future Directions}
\label{limits}

While \textsf{\textbf{\textsc{VLDBench}}} advances multimodal disinformation research, it inherits limitations typical of empirical studies and highlights opportunities for improvement.

\paragraph{Sampling bias}
Reliance on pre-verified news sources may underrepresent strategies prevalent on fringe or lightly regulated platforms (e.g., social media), potentially limiting adaptability to rapidly evolving tactics. Future releases should incorporate decentralized and user-generated content to improve generalizability.

\paragraph{AI-assisted annotation}
LLM-assisted (multimodal) annotations can reproduce model biases or behave unpredictably. Although we employed mitigation techniques, hyperparameter tuning, multi-pass labeling, and human verification, the probabilistic nature of these systems raises fairness concerns. Misclassifications in sensitive contexts can disproportionately affect marginalized communities \cite{raza2024developing}, underscoring the importance of adversarial testing and dedicated bias-detection frameworks.

\paragraph{Effect of image and cross-modal perturbations}
Isolated image perturbations (Gaussian blur, additive noise, resizing) yield moderate performance declines in LVLMs, typically \mbox{2--3\%} macro-F1, indicating reasonable stability when only the image channel is degraded (see Table~\ref{tab:perturbations1}). In contrast, cross-modal mismatches and combined text+image attacks produce substantially larger \emph{degradations} (often $>$10\% macro-F1; see Table~\ref{tab:adversarial_performance}), revealing a key vulnerability in multimodal reasoning pipelines when distortions span both channels.

\paragraph{Scope and generalizability}
 \textsf{\textbf{\textsc{VLDBench}}} targets the \emph{news domain}: mainstream, English-language articles paired with editorial images (2023–2024). Results should be interpreted as evidence of performance in newsroom-style contexts, not as claims about unregulated or multilingual platforms. Our external checks (r/Fakeddit; Table~\ref{tab:ood}) are stress tests rather than scope extensions: they rely on falsity-based proxies and social-media distributions that differ from our intent-labeled news setting. Tactics common on fringe or non-English platforms (e.g., memes, OCR-heavy screenshots, code-switching) are under-represented. Extending coverage to user-generated and multilingual sources is a key direction for future releases .

\paragraph{Modeling choices and compute}
To maintain comparability, we evaluate baseline models and IFT variants; we do not explore more advanced strategies (e.g., self-feedback, critique modules, visual program synthesis), which remain promising for future work. Finally, the compute demands of training/evaluating large multimodal models can constrain participation. Developing lightweight models and decentralized training (e.g., federated learning) can help democratize access.

Despite these limitations, \textsf{\textbf{\textsc{VLDBench}}} provides a solid foundation for benchmarking multimodal disinformation detection, enabling the community to refine methods, expand coverage, and address ethical, technical, and social risks.

\subsection{Responsible Use}
Responsible use is a foundational principle in the design of \textsf{\textbf{\textsc{VLDBench}}}. The dataset is ethically curated, excludes personal data, and is dedicated solely to research on the detection of disinformation. While some examples necessarily include offensive or harmful content (inherent to the phenomenon being studied), we document known risks (e.g., stereotyping, inflammatory language) in accompanying datasheets. Users must comply with usage terms that prohibit harmful applications, including censorship, surveillance, or the generation of targeted disinformation. Although no dataset can fully prevent misuse, we promote accountability through transparency about annotation protocols, model training, and evaluation procedures, supporting reproducibility and independent scrutiny. We envision \textsf{\textbf{\textsc{VLDBench}}} as a foundation for media-literacy initiatives, fact-checking and unbiased newsroom workflows, and policy discussions on AI governance. By prioritizing ethical design and equitable access, our work aims to empower communities and institutions to combat disinformation while fostering trust in digital ecosystems.

\section{Conclusion}

\textsf{\textbf{\textsc{VLDBench}}} addresses the urgent challenge of disinformation in the generative-AI era through responsible data stewardship and governance alignment (e.g., EU AI Act, NIST AI RMF, MIT AI Risk Repository). To our knowledge, it is the first \emph{human-verified}, multimodal disinformation detection benchmark that evaluates modern LLMs and VLMs at scale, comprising \textbf{62{,}678} instances (31{,}339 text-only; 31{,}339 text–image pairs) across \textbf{13} topical categories. While compatible with standard ML pipelines, its design is tailored to the risks introduced by multimodal generative models.
Empirically, we find that (i) VLMs outperform strong text-only baselines; (ii) IFT yields consistent gains; (iii) cross-modal and combined perturbations induce the largest performance drops; (iv) scaling within families improves accuracy and F1; and (v) out-of-distribution evaluation reveals meaningful domain-shift effects. These results offer a rigorous, reproducible basis for stress-testing models against realistic deception strategies.

Some key constraints include: (1) reliance on pre-verified news sources, which may underrepresent tactics prevalent on lightly regulated platforms; (2) hybrid AI–human annotation that can inherit biases from both sources~\cite{gilardi2023chatgpt}; (3) compute constraints that limited the breadth of model-scale ablations; (4) English-only content, limiting cross-lingual generalizability; and (5) exclusive evaluation of open-source models for transparency and accessibility, while a closed-source model (GPT-4o) was used solely for annotation and automated judging and is excluded from the leaderboard. These factors motivate future releases that broaden source coverage (user-generated and multilingual content), incorporate robustness-aware training, and expand fairness and calibration analyses.
Despite these limitations, \textsf{\textbf{\textsc{VLDBench}}} provides a critical step toward systematic, governance-aligned benchmarking of disinformation detection. We hope it enables the community to advance safer, more reliable language and vision systems and to inform evidence-based policy and media-literacy efforts.

\bibliographystyle{elsarticle-num}
\bibliography{references}

@article{farooq2025evaluating,
  title={Evaluating and Regulating Agentic AI: A Study of Benchmarks, Metrics, and Regulation},
  author={Farooq, Azib and Raza, Shaina and Karim, Md Nazmul and Iqbal, Hasan and Vasilakos, Athanasios V and Emmanouilidis, Christos},
  journal={Metrics, and Regulation},
  year={2025}
}

@misc{pai2024,
  author = {Partnership on AI},
  title = {Framework for Responsible AI and Synthetic Media},
  year = {2024},
  url = {https://partnershiponai.org/},
  note = {Accessed: May 30, 2025}
}

@article{luo2021newsclippings,
  title={Newsclippings: Automatic generation of out-of-context multimodal media},
  author={Luo, Grace and Darrell, Trevor and Rohrbach, Anna},
  journal={arXiv preprint arXiv:2104.05893},
  year={2021}
}

@inproceedings{li2020celeb,
  title={Celeb-df: A large-scale challenging dataset for deepfake forensics},
  author={Li, Yuezun and Yang, Xin and Sun, Pu and Qi, Honggang and Lyu, Siwei},
  booktitle={Proceedings of the IEEE/CVF conference on computer vision and pattern recognition},
  pages={3207--3216},
  year={2020}
}

@misc{c2pa2024,
  author = {{C2PA}},
  title = {Coalition for Content Provenance and Authenticity (C2PA)},
  year = {2024},
  url = {https://c2pa.org/},
  note = {Accessed: May 30, 2025}
}

@misc{dolhansky2020dfdc,
  title         = {The Deepfake Detection Challenge (DFDC) Dataset},
  author        = {Dolhansky, Brian and Howes, Joanna and Pflaum, Ben and Baram, Nicole and Ferrer, Cristian Canton and and others},
  year          = {2020},
  eprint        = {2006.07397},
  archivePrefix = {arXiv},
  primaryClass  = {cs.CV}
}

@inproceedings{roessler2019faceforensicspp,
  title={Faceforensics++: Learning to detect manipulated facial images},
  author={Rossler, Andreas and Cozzolino, Davide and Verdoliva, Luisa and Riess, Christian and Thies, Justus and Nie{\ss}ner, Matthias},
  booktitle={Proceedings of the IEEE/CVF international conference on computer vision},
  pages={1--11},
  year={2019}
}

@misc{techaccord2024,
  author = {{Microsoft}},
  title = {Tech Accord to Combat Deceptive Use of AI in 2024 Elections},
  year = {2024},
  month = {February 16},
  url = {https://blogs.microsoft.com/on-the-issues/2024/02/16/ai-deepfakes-elections-munich-tech-accord/},
  note = {Accessed: May 30, 2025}
}

@article{zhang2024multitrust,
  title={Multitrust: A comprehensive benchmark towards trustworthy multimodal large language models},
  author={Zhang, Yichi and Huang, Yao and Sun, Yitong and Liu, Chang and Zhao, Zhe and Fang, Zhengwei and Wang, Yifan and Chen, Huanran and Yang, Xiao and Wei, Xingxing and others},
  journal={Advances in Neural Information Processing Systems},
  volume={37},
  pages={49279--49383},
  year={2024}
}

@article{shao2024dgm4++,
  title={Detecting and Grounding Multi-Modal Media Manipulation and Beyond},
  author={Shao, Rui and Wu, Tianxing and Wu, Jianlong and Nie, Liqiang and Liu, Ziwei},
  journal={IEEE Transactions on Pattern Analysis and Machine Intelligence (TPAMI)},
  year={2024},
}

@misc{cai2024internlm2,
      title={InternLM2 Technical Report},
        author={Zheng Cai et al.},
      year={2024},
      eprint={2403.17297},
      archivePrefix={arXiv},
      primaryClass={cs.CL}
}

@article{li2024towards,
  title={Towards Multimodal Disinformation Detection by Vision-language Knowledge Interaction},
  author={Li, Qilei and Gao, Mingliang and Zhang, Guisheng and Zhai, Wenzhe and Chen, Jinyong and Jeon, Gwanggil},
  journal={Information Fusion},
  volume={102},
  pages={102037},
  year={2024},
  publisher={Elsevier}
}

@article{wu2024deepseek,
  title={Deepseek-vl2: Mixture-of-experts vision-language models for advanced multimodal understanding},
  author={Wu, Zhiyu and Chen, Xiaokang and Pan, Zizheng and Liu, Xingchao and Liu, Wen and Dai, Damai and Gao, Huazuo and Ma, Yiyang and Wu, Chengyue and Wang, Bingxuan and others},
  journal={arXiv preprint arXiv:2412.10302},
  year={2024}
}

@article{liu2024deepseek,
  title={Deepseek-v2: A strong, economical, and efficient mixture-of-experts language model},
  author={Liu, Aixin and Feng, Bei and Wang, Bin and Wang, Bingxuan and Liu, Bo and Zhao, Chenggang and Dengr, Chengqi and Ruan, Chong and Dai, Damai and Guo, Daya and others},
  journal={arXiv preprint arXiv:2405.04434},
  year={2024}
}

@article{zellers2019defending,
  title={Defending against neural fake news},
  author={Zellers, Rowan and Holtzman, Ari and Rashkin, Hannah and Bisk, Yonatan and Farhadi, Ali and Roesner, Franziska and Choi, Yejin},
  journal={Advances in neural information processing systems},
  volume={32},
  year={2019}
}

@misc{phi3vision2024,
  author       = {Microsoft},
  title        = {Phi-3-Vision-128K-Instruct},
  year         = {2024},
  publisher    = {Hugging Face},
  journal      = {Hugging Face Repository},
  url          = {https://huggingface.co/microsoft/Phi-3-vision-128k-instruct},
  note         = {A lightweight, state-of-the-art open multimodal model supporting up to 128,000 tokens of context, integrating vision and language processing capabilities.},
}

@misc{phi3mini2024,
  author       = {Microsoft Research},
  title        = {Phi-3-Mini-128K-Instruct},
  year         = {2024},
  publisher    = {Hugging Face},
  journal      = {Hugging Face Repository},
  url          = {https://huggingface.co/microsoft/Phi-3-mini-128k-instruct},
  note         = {Dense 3.8B parameter, decoder-only Transformer model with supervised fine-tuning and alignment techniques, supporting up to 128,000 tokens of context.},
}

@inproceedings{chen2024internvl,
  title={Internvl: Scaling up vision foundation models and aligning for generic visual-linguistic tasks},
  author={Chen, Zhe and Wu, Jiannan and Wang, Wenhai and Su, Weijie and Chen, Guo and Xing, Sen and Zhong, Muyan and Zhang, Qinglong and Zhu, Xizhou and Lu, Lewei and others},
  booktitle={Proceedings of the IEEE/CVF Conference on Computer Vision and Pattern Recognition},
  pages={24185--24198},
  year={2024}
}

@article{Qwen2VL,
  title={Qwen2-VL: Enhancing Vision-Language Model's Perception of the World at Any Resolution},
  author={Wang, Peng and Bai, Shuai and Tan, Sinan and Wang, Shijie and Fan, Zhihao and Bai, Jinze and Chen, Keqin and Liu, Xuejing and Wang, Jialin and Ge, Wenbin and Fan, Yang and Dang, Kai and Du, Mengfei and Ren, Xuancheng and Men, Rui and Liu, Dayiheng and Zhou, Chang and Zhou, Jingren and Lin, Junyang},
  journal={arXiv preprint arXiv:2409.12191},
  year={2024}
}

@misc{morris2020textattack,
    title={TextAttack: A Framework for Adversarial Attacks, Data Augmentation, and Adversarial Training in NLP},
    author={John X. Morris and Eli Lifland and Jin Yong Yoo and Jake Grigsby and Di Jin and Yanjun Qi},
    year={2020},
    eprint={2005.05909},
    archivePrefix={arXiv},
    primaryClass={cs.CL}
}

@misc{EUCodeDisinfo,
  title        = {Code of Practice on Disinformation},
  author       = {{European Commission}},
  year         = 2018,
  url          = {https://digital-strategy.ec.europa.eu/en/policies/code-practice-disinformation},
  note         = {Accessed: 2024-11-18}
}

@misc{EUDemocracyActionPlan,
  title        = {European Democracy Action Plan},
  author       = {{European Commission}},
  year         = 2020,
  url          = {https://ec.europa.eu/commission/presscorner/detail/en/ip_20_2250},
  note         = {Accessed: 2024-11-18}
}

@misc{DigitalServicesAct,
  title        = {Regulation (EU) 2022/2065 of the European Parliament and of the Council of 19 October 2022 on a Single Market for Digital Services (Digital Services Act)},
  author       = {{European Union}},
  year         = 2022,
  url          = {https://eur-lex.europa.eu/legal-content/EN/TXT/?uri=CELEX\%3A32022R2065},
  note         = {Accessed: 2024-11-18}
}

@misc{zheng2023judging,
      title={Judging LLM-as-a-judge with MT-Bench and Chatbot Arena},
      author={Lianmin Zheng and Wei-Lin Chiang and Ying Sheng and Siyuan Zhuang and Zhanghao Wu and Yonghao Zhuang and Zi Lin and Zhuohan Li and Dacheng Li and Eric. P Xing and Hao Zhang and Joseph E. Gonzalez and Ion Stoica},
      year={2023},
      eprint={2306.05685},
      archivePrefix={arXiv},
      primaryClass={cs.CL}
}

@article{wan2025evaluating,
  title={Evaluating Sex and Age Biases in Multimodal Large Language Models for Skin Disease Identification from Dermatoscopic Images},
  author={Wan, Zhiyu and Guo, Yuhang and Bao, Shunxing and Wang, Qian and Malin, Bradley A},
  journal={Health Data Science},
  volume={5},
  pages={0256},
  year={2025},
  publisher={AAAS}
}

@regulation{EU_AI_Act_2024,
  title        = {Regulation (EU) 2024/1689 of the European Parliament and of the Council of 13 June 2024 laying down harmonised rules on artificial intelligence (Artificial Intelligence Act)},
  institution  = {European Parliament and Council of the European Union},
  type         = {Regulation},
  number       = {2024/1689},
  date         = {2024-06-13},
  journal       = {Official Journal of the European Union, L},
  pages         = {1--81},  
  note          = {Entered into force 1 August 2024; full implementation phased in by 2026.},
  url           = {https://eur-lex.europa.eu/eli/reg/2024/1689/oj/eng}
}

@misc{Canada_Online_News_Act_2023,
  author       = {Government of Canada},
  title        = {Online News Act (Bill C-18): An Act respecting online communications platforms that make news content available to persons in Canada},
  howpublished = {\url{https://www.justice.gc.ca/eng/csj-sjc/pl/charter-charte/c18_1.html}},
  note         = {Received Royal Assent 22 June 2023; regulates digital news intermediaries to enhance fairness for Canadian news businesses.},
  year         = {2023}
}

@misc{CounteringForeignPropaganda,
  title        = {Countering Foreign Propaganda and Disinformation Act},
  author       = {{U.S. Congress}},
  year         = 2016,
  url          = {https://www.congress.gov/bill/114th-congress/house-bill/5181},
  note         = {Accessed: 2024-11-18}
}

@misc{UK_Online_Safety_Act_2023,
  author       = {UK Parliament},
  title        = {Online Safety Act 2023 (c. 50)},
  howpublished = {\url{https://www.legislation.gov.uk/ukpga/2023/50}},
  note         = {An Act to make provision about online content that is harmful or illegal, duties for services that host or search for such content, and for connected purposes. Received Royal Assent 26 October 2023; is in force in part as of September 2025.},
  year         = {2023}
}

@misc{FCC_AIRobocalls_2024,
  title   = {FCC Declaratory Ruling: AI-Generated Voices in Robocalls are ``Artificial'' under the TCPA},
  author  = {{Federal Communications Commission}},
  year    = {2024},
  howpublished = {\url{https://www.fcc.gov/document/fcc-makes-ai-generated-voices-robocalls-illegal}},
  urldate = {2025-09-21},
  note    = {FCC 24-17; adopted Feb.\ 8, 2024}
}

@misc{MaliciousDeepFakeProhibition,
  title        = {Malicious Deep Fake Prohibition Act},
  author       = {{U.S. Congress}},
  year         = 2018,
  url          = {https://www.congress.gov/bill/115th-congress/senate-bill/3805},
  note         = {Accessed: 2024-11-18}
}

@misc{OnlineSafetyActUK,
  title        = {Online Safety Act},
  author       = {{UK Government}},
  year         = 2021,
  url          = {https://www.gov.uk/government/collections/online-safety-bill},
  note         = {Accessed: 2024-11-18}
}

@misc{ChinaDeepSynthesisProvision,
  title        = {Provision on the Administration of Deep Synthesis Internet Information Service},
  author       = {{Cyberspace Administration of China}},
  year         = 2022,
  url          = {http://www.cac.gov.cn/2022-01/28/c_1644970458520968.htm},
  note         = {Accessed: 2024-11-18}
}

@misc{anthropic2023core,
  title={Core Views on AI Safety: When, Why, What, and How},
  author={Anthropic, PBC},
  year={2023},
  publisher={Anthropic. https://www. anthropic. com/index/core-views-on-ai-safety}
}

@article{gruppi2021nela,
  title={NELA-GT-2020: A large multi-labelled news dataset for the study of misinformation in news articles},
  author={Gruppi, Maur{\'\i}cio and Horne, Benjamin D and Adal{\i}, Sibel},
  journal={arXiv preprint arXiv:2102.04567},
  year={2021}
}

@misc{allsides_mediabiaschart,
  author = {{AllSides}},
  title = {Media Bias Chart},
  year = {2024},
  url = {https://www.allsides.com/media-bias/media-bias-chart},
  note = {Accessed: 2024-02-13}
}

@article{liu2024fakenewsgpt4,
  title={FakeNewsGPT4: Advancing Multimodal Fake News Detection through Knowledge-Augmented LVLMs},
  author={Liu, Xuannan and Li, Peipei and Huang, Huaibo and Li, Zekun and Cui, Xing and Liang, Jiahao and Qin, Lixiong and Deng, Weihong and He, Zhaofeng},
  journal={arXiv preprint arXiv:2403.01988},
  year={2024}
}

@article{xuan2024lemma,
  title={LEMMA: Towards LVLM-Enhanced Multimodal Misinformation Detection with External Knowledge Augmentation},
  author={Xuan, Keyang and Yi, Li and Yang, Fan and Wu, Ruochen and Fung, Yi R and Ji, Heng},
  journal={arXiv preprint arXiv:2402.11943},
  year={2024}
}

@article{song2021multimodal,
  title={A multimodal fake news detection model based on crossmodal attention residual and multichannel convolutional neural networks},
  author={Song, Chenguang and Ning, Nianwen and Zhang, Yunlei and Wu, Bin},
  journal={Information Processing \& Management},
  volume={58},
  number={1},
  pages={102437},
  year={2021},
  publisher={Elsevier}
}

@article{segura2022multimodal,
  title={Multimodal fake news detection},
  author={Segura-Bedmar, Isabel and Alonso-Bartolome, Santiago},
  journal={Information},
  volume={13},
  number={6},
  pages={284},
  year={2022},
  publisher={MDPI}
}

@inproceedings{wu2021multimodal,
  title={Multimodal fusion with co-attention networks for fake news detection},
  author={Wu, Yang and Zhan, Pengwei and Zhang, Yunjian and Wang, Liming and Xu, Zhen},
  booktitle={Findings of the association for computational linguistics: ACL-IJCNLP 2021},
  pages={2560--2569},
  year={2021}
}

@article{zong2024safety,
  title={Safety fine-tuning at (almost) no cost: A baseline for vision large language models},
  author={Zong, Yongshuo and Bohdal, Ondrej and Yu, Tingyang and Yang, Yongxin and Hospedales, Timothy},
  journal={arXiv preprint arXiv:2402.02207},
  year={2024}
}

@article{zhang2023safetybench,
  title={Safetybench: Evaluating the safety of large language models with multiple choice questions},
  author={Zhang, Zhexin and Lei, Leqi and Wu, Lindong and Sun, Rui and Huang, Yongkang and Long, Chong and Liu, Xiao and Lei, Xuanyu and Tang, Jie and Huang, Minlie},
  journal={arXiv preprint arXiv:2309.07045},
  year={2023}
}

@article{slattery_2024ai,
  title={The ai risk repository: A comprehensive meta-review, database, and taxonomy of risks from artificial intelligence},
  author={Slattery, Peter and Saeri, Alexander K and Grundy, Emily AC and Graham, Jess and Noetel, Michael and Uuk, Risto and Dao, James and Pour, Soroush and Casper, Stephen and Thompson, Neil},
  journal={arXiv preprint arXiv:2408.12622},
  year={2024}
}

@techreport{EUActionPlan2018,
  title = {Action Plan against Disinformation},
  author = {{European Commission} and {High Representative of the Union for Foreign Affairs and Security Policy}},
  year = {2018},
  month = {December},
  day = {5},
  institution = {European Commission},
  type = {Joint Communication},
  number = {JOIN(2018) 36 final},
  address = {Brussels},
  url = {https://www.eeas.europa.eu/sites/default/files/action_plan_against_disinformation.pdf}
}

@misc{CPA2024Handbook,
  author = {{Commonwealth Parliamentary Association}},
  title = {Handbook on Disinformation, AI, and Synthetic Media},
  year = {2024},
  howpublished = {\url{https://www.cpahq.org/media/sphl0rft/handbook-on-disinformation-ai-and-synthetic-media.pdf}},
  note = {Accessed: 2024-10-15}
}

@article{yang2024qwen2,
  title={Qwen2 technical report},
  author={Yang, An and Yang, Baosong and Hui, Binyuan and Zheng, Bo and Yu, Bowen and Zhou, Chang and Li, Chengpeng and Li, Chengyuan and Liu, Dayiheng and Huang, Fei and others},
  journal={arXiv preprint arXiv:2407.10671},
  year={2024}
}

@misc{benkler2018network,
  title={Network propaganda: manipulation, disinformation, and radicalization in American Politics},
  author={Benkler, Y},
  year={2018},
  publisher={Oxford University Press}
}

@article{chen2023combating,
  title={Combating misinformation in the age of llms: Opportunities and challenges},
  author={Chen, Canyu and Shu, Kai},
  journal={AI Magazine},
  year={2023},
  publisher={Wiley Online Library}
}

@misc{EUAIAct2024,
  title = {Artificial {Intelligence} {Act}, {Article} 14: {Human} oversight},
  author = {{European Union}},
  year = {2024},
  url = {https://artificialintelligenceact.eu/article/14/},
  note = {Accessed on November 08, 2024},
  organization = {European Union}
}

@inproceedings{khattar2019mvae,
  title={Mvae: Multimodal variational autoencoder for fake news detection},
  author={Khattar, Dhruv and Goud, Jaipal Singh and Gupta, Manish and Varma, Vasudeva},
  booktitle={The world wide web conference},
  pages={2915--2921},
  year={2019}
}

@article{bahad2019fake,
  title={Fake news detection using bi-directional LSTM-recurrent neural network},
  author={Bahad, Pritika and Saxena, Preeti and Kamal, Raj},
  journal={Procedia Computer Science},
  volume={165},
  pages={74--82},
  year={2019},
  publisher={Elsevier}
}

@inproceedings{rubin2016fake,
  title={Fake news or truth? using satirical cues to detect potentially misleading news},
  author={Rubin, Victoria L and Conroy, Niall and Chen, Yimin and Cornwell, Sarah},
  booktitle={Proceedings of the second workshop on computational approaches to deception detection},
  pages={7--17},
  year={2016}
}

@techreport{bontcheva2024generative,
  title={Generative AI and Disinformation: Recent Advances, Challenges, and Opportunities},
  author={Bontcheva, Kalina and Papadopoulous, Symeon and Tsalakanidou, Filareti and Gallotti, Riccardo and Dutkiewicz, Lidia and Krack, No{\'e}mie and Teyssou, Denis and Severio Nucci, Francesco and Spangenberg, Jochen and Srba, Ivan and others},
  institution="University of Dubrovnik",
  year={2024}
}

@misc{samvelyan2024rainbowteaming,
      title={Rainbow Teaming: Open-Ended Generation of Diverse Adversarial Prompts}, 
      author={Mikayel Samvelyan and Sharath Chandra Raparthy and Andrei Lupu and Eric Hambro and Aram H. Markosyan and Manish Bhatt and Yuning Mao and Minqi Jiang and Jack Parker-Holder and Jakob Foerster and Tim Rocktäschel and Roberta Raileanu},
      year={2024},
      journal={arXiv journal arXiv: 2402.16822}
}

@misc{vidgen2024introducingv05aisafety,
  title={Introducing v0.5 of the AI Safety Benchmark from MLCommons},
  author={Bertie Vidgen et al.},
  year={2024},
  journal={arXiv:2404.12241}
}

@misc{OnlineNewsAct2023,
  title        = {Online News Act},
  author       = {Parliament of Canada},
  year         = 2023,
  url          = {https://www.parl.ca/DocumentViewer/en/44-1/bill/C-18/royal-assent},
  note         = {Accessed: 2024-11-18}
}

@misc{govCanada_disinformation,
  title = {Combating Disinformation},
  author = {{Global Affairs Canada}},
  year = {2023},
  url = {https://www.international.gc.ca/world-monde/issues_development-enjeux_developpement/peace_security-paix_securite/combatt-disinformation-desinformation.aspx?lang=eng},
  note = {Accessed: 2024-11-08},
  publisher = {Government of Canada}
}

@techreport{EuropeanCommission2020,
  title = {European Democracy Action Plan: making EU democracies stronger},
  author = {{European Commission}},
  year = {2020},
  month = {December},
  day = {3},
  institution = {European Commission},
  type = {Press Release},
  number = {IP/20/2250},
  address = {Brussels},
  url = {https://ec.europa.eu/commission/presscorner/detail/en/ip_20_2250}
}

@article{shu2017fake,
  title={Fake news detection on social media: A data mining perspective},
  author={Shu, Kai and Sliva, Amy and Wang, Suhang and Tang, Jiliang and Liu, Huan},
  journal={ACM SIGKDD explorations newsletter},
  volume={19},
  number={1},
  pages={22--36},
  year={2017},
  publisher={ACM New York, NY, USA}
}

@article{santos2021misinformation,
  title={Misinformation, disinformation, and malinformation: clarifying the definitions and examples in disinfodemic times},
  author={Santos-D’Amorim, Karen and de Oliveira Miranda, M{\'a}jory K Fernandes},
  journal={Encontros Bibli: revista eletr{\^o}nica de biblioteconomia e ci{\^e}ncia da informa{\c{c}}{\~a}o},
  volume={26},
  year={2021},
  publisher={Universidade Federal de Santa Catarina}
}

@misc{canada_aia,
  title = {Algorithmic Impact Assessment tool},
  author = {{Government of Canada}},
  year = {2024},
  month = {5},
  day = {30},
  url = {https://www.canada.ca/en/government/system/digital-government/digital-government-innovations/responsible-use-ai/algorithmic-impact-assessment.html},
  organization = {Government of Canada},
  note = {Accessed: September 13, 2024}
}

@article{kim2024meganno+,
  title={Meganno+: A human-llm collaborative annotation system},
  author={Kim, Hannah and Mitra, Kushan and Chen, Rafael Li and Rahman, Sajjadur and Zhang, Dan},
  journal={arXiv preprint arXiv:2402.18050},
  year={2024}
}

@article{gilardi2023chatgpt,
  title={ChatGPT outperforms crowd workers for text-annotation tasks},
  author={Gilardi, Fabrizio and Alizadeh, Meysam and Kubli, Ma{\"e}l},
  journal={Proceedings of the National Academy of Sciences},
  volume={120},
  number={30},
  pages={e2305016120},
  year={2023},
  publisher={National Acad Sciences}
}

@inproceedings{papado2023misinformer,
author = {Papadopoulos, Stefanos-Iordanis and Koutlis, Christos and Papadopoulos, Symeon and Petrantonakis, Panagiotis},
title = {Synthetic Misinformers: Generating and Combating Multimodal Misinformation},
year = {2023},
isbn = {9798400701870},
publisher = {Association for Computing Machinery},
address = {New York, NY, USA},
url = {https://doi.org/10.1145/3592572.3592842},
doi = {10.1145/3592572.3592842},
booktitle = {Proceedings of the 2nd ACM International Workshop on Multimedia AI against Disinformation},
pages = {36–44},
numpages = {9},
location = {Thessaloniki, Greece},
series = {MAD '23}
}

@InProceedings{Qi_2024_CVPR,
    author    = {Qi, Peng and Yan, Zehong and Hsu, Wynne and Lee, Mong Li},
    title     = {SNIFFER: Multimodal Large Language Model for Explainable Out-of-Context Misinformation Detection},
    booktitle = {Proceedings of the IEEE/CVF Conference on Computer Vision and Pattern Recognition (CVPR)},
    month     = {June},
    year      = {2024},
    pages     = {13052-13062}
}

@techreport{meta2024llama,
    author = {AI at Meta},
    title = {Llama 3.2: Revolutionizing edge AI and vision with open, customizable models} ,
    institution = {Meta},
    year = 2024,
link = {https://ai.meta.com/blog/llama-3-2-connect-2024-vision-edge-mobile-devices/}
}

@techreport{meta2024llama31,
    author = {AI at Meta},
    title = {Introducing Llama 3.1: Our most capable models to date} ,
    institution = {Meta},
    year = 2024,
link = {https://ai.meta.com/blog/meta-llama-3-1/}
}

@techreport{mistral20237b,
    author = {Mistral AI Team} ,
    title = {Mistral 7B
The best 7B model to date},
    institution = {Mistral} ,
    year = {2023},
link = {https://mistral.ai/news/announcing-mistral-7b/}
}

@techreport{mistral2024pix,
    author = {Mistral AI Team},
    title = {Pixtral 12B - the first-ever multimodal Mistral model},
    institution = {Mistral},
    year = 2024
}

@inproceedings{NEURIPS2023LLaVA,
 author = {Liu, Haotian and Li, Chunyuan and Wu, Qingyang and Lee, Yong Jae},
 booktitle = {Advances in Neural Information Processing Systems},
 editor = {A. Oh and T. Naumann and A. Globerson and K. Saenko and M. Hardt and S. Levine},
 pages = {34892--34916},
 publisher = {Curran Associates, Inc.},
 title = {Visual Instruction Tuning},
 url = {https://proceedings.neurips.cc/paper_files/paper/2023/file/6dcf277ea32ce3288914faf369fe6de0-Paper-Conference.pdf},
 volume = {36},
 year = {2023}
}

@inproceedings{mumin2022dataset,
author = {Nielsen, Dan S. and McConville, Ryan},
title = {MuMiN: A Large-Scale Multilingual Multimodal Fact-Checked Misinformation Social Network Dataset},
year = {2022},
isbn = {9781450387323},
publisher = {Association for Computing Machinery},
address = {New York, NY, USA},
url = {https://doi.org/10.1145/3477495.3531744},
doi = {10.1145/3477495.3531744},
booktitle = {Proceedings of the 45th International ACM SIGIR Conference on Research and Development in Information Retrieval},
pages = {3141–3153},
numpages = {13},
location = {Madrid, Spain},
series = {SIGIR '22}
}

@inproceedings{nakamura-etal-2020-fakeddit,
    title = "{F}akeddit: A New Multimodal Benchmark Dataset for Fine-grained Fake News Detection",
    author = "Nakamura, Kai  and
      Levy, Sharon  and
      Wang, William Yang",
    editor = "Calzolari, Nicoletta  and
      B{\'e}chet, Fr{\'e}d{\'e}ric  and
      Blache, Philippe  and
      Choukri, Khalid  and
      Cieri, Christopher  and
      Declerck, Thierry  and
      Goggi, Sara  and
      Isahara, Hitoshi  and
      Maegaard, Bente  and
      Mariani, Joseph  and
      Mazo, H{\'e}l{\`e}ne  and
      Moreno, Asuncion  and
      Odijk, Jan  and
      Piperidis, Stelios",
    booktitle = "Proceedings of the Twelfth Language Resources and Evaluation Conference",
    month = may,
    year = "2020",
    address = "Marseille, France",
    publisher = "European Language Resources Association",
    url = "https://aclanthology.org/2020.lrec-1.755",
    pages = "6149--6157",
    language = "English",
    ISBN = "979-10-95546-34-4",
}

@inproceedings{ NewsBag2020dataset,
      title={ NewsBag: A multimodal benchmark dataset for fake news detection },
      author={  S.  Jindal  and   R.  Sood  and   Richa  Singh  and   Mayank  Vatsa  and   T.  Chakraborty   },
      booktitle ={ CEUR Workshop Proceedings },
      year={ 2020 },
      publisher={ CEUR-WS },
    volume={ 2560 },
  pages={ 138 -- 145  },
      issn={ 16130073 }
}

@misc{suryavardan2023factify2multimodalfake,
      title={Factify 2: A Multimodal Fake News and Satire News Dataset}, 
      author={S Suryavardan and Shreyash Mishra and Parth Patwa and Megha Chakraborty and Anku Rani and Aishwarya Reganti and Aman Chadha and Amitava Das and Amit Sheth and Manoj Chinnakotla and Asif Ekbal and Srijan Kumar},
      year={2023},
      eprint={2304.03897},
      archivePrefix={arXiv},
      primaryClass={cs.CL},
      url={https://arxiv.org/abs/2304.03897}, 
}

@inproceedings{wang-2017-liar,
    title = "{``}Liar, Liar Pants on Fire{''}: A New Benchmark Dataset for Fake News Detection",
    author = "Wang, William Yang",
    editor = "Barzilay, Regina  and
      Kan, Min-Yen",
    booktitle = "Proceedings of the 55th Annual Meeting of the Association for Computational Linguistics (Volume 2: Short Papers)",
    month = jul,
    year = "2017",
    address = "Vancouver, Canada",
    publisher = "Association for Computational Linguistics",
    url = "https://aclanthology.org/P17-2067",
    doi = "10.18653/v1/P17-2067",
    pages = "422--426"
}

@misc{thorne2018feverlargescaledatasetfact,
      title={FEVER: a large-scale dataset for Fact Extraction and VERification}, 
      author={James Thorne and Andreas Vlachos and Christos Christodoulopoulos and Arpit Mittal},
      year={2018},
      eprint={1803.05355},
      archivePrefix={arXiv},
      primaryClass={cs.CL},
      url={https://arxiv.org/abs/1803.05355}, 
}

@article{Santia_Williams_2018, title={BuzzFace: A News Veracity Dataset with Facebook User Commentary and Egos}, volume={12}, url={https://ojs.aaai.org/index.php/ICWSM/article/view/14985}, DOI={10.1609/icwsm.v12i1.14985}, number={1}, journal={Proceedings of the International AAAI Conference on Web and Social Media}, author={Santia, Giovanni and Williams, Jake}, year={2018}, month={Jun.}, pages={531-540} }

@article{raza2024developing,
  title={Developing Safe and Responsible Large Language Model: Can We Balance Bias Reduction and Language Understanding in Large Language Models?},
  author={Raza, Shaina and Bamgbose, Oluwanifemi and Ghuge, Shardul and Tavakol, Fatemeh and Reji, Deepak John and Bashir, Syed Raza},
  journal={arXiv preprint arXiv:2404.01399},
  year={2024}
}

@article{gpt4o,
  title={Gpt-4o system card},
  author={Hurst, Aaron and Lerer, Adam and Goucher, Adam P and Perelman, Adam and Ramesh, Aditya and Clark, Aidan and Ostrow, AJ and Welihinda, Akila and Hayes, Alan and Radford, Alec and others},
  journal={arXiv preprint arXiv:2410.21276},
  year={2024}
}

@article{vayani2024all,
  title={All languages matter: Evaluating lmms on culturally diverse 100 languages},
  author={Vayani, Ashmal and Dissanayake, Dinura and Watawana, Hasindri and Ahsan, Noor and Sasikumar, Nevasini and Thawakar, Omkar and Ademtew, Henok Biadglign and Hmaiti, Yahya and Kumar, Amandeep and Kuckreja, Kartik and others},
  journal={arXiv preprint arXiv:2411.16508},
  year={2024}
}

@article{chen2025janus,
  title={Janus-Pro: Unified Multimodal Understanding and Generation with Data and Model Scaling},
  author={Chen, Xiaokang and Wu, Zhiyu and Liu, Xingchao and Pan, Zizheng and Liu, Wen and Xie, Zhenda and Yu, Xingkai and Ruan, Chong},
  journal={arXiv preprint arXiv:2501.17811},
  year={2025}
}

@misc{glm2024chatglm,
      title={ChatGLM: A Family of Large Language Models from GLM-130B to GLM-4 All Tools}, 
      author={Team GLM},
      year={2024},
      eprint={2406.12793},
      archivePrefix={arXiv},
      primaryClass={id='cs.CL' full_name='Computation and Language' is_active=True alt_name='cmp-lg' in_archive='cs' is_general=False description='Covers natural language processing. Roughly includes material in ACM Subject Class I.2.7. Note that work on artificial languages (programming languages, logics, formal systems) that does not explicitly address natural-language issues broadly construed (natural-language processing, computational linguistics, speech, text retrieval, etc.) is not appropriate for this area.'}
}

@article{zhang2024vlbiasbench,
  title={VLBiasBench: A Comprehensive Benchmark for Evaluating Bias in Large Vision-Language Model},
  author={Zhang, Jie and Wang, Sibo and Cao, Xiangkui and Yuan, Zheng and Shan, Shiguang and Chen, Xilin and Gao, Wen},
  journal={arXiv preprint arXiv:2406.14194},
  year={2024}
}

@inproceedings{liu2025mm,
  title={Mm-safetybench: A benchmark for safety evaluation of multimodal large language models},
  author={Liu, Xin and Zhu, Yichen and Gu, Jindong and Lan, Yunshi and Yang, Chao and Qiao, Yu},
  booktitle={European Conference on Computer Vision},
  pages={386--403},
  year={2025},
  organization={Springer}
}

@article{shu_fakenewsnet_2020,
    title = {{FakeNewsNet}: {A} {Data} {Repository} with {News} {Content}, {Social} {Context}, and {Spatiotemporal} {Information} for {Studying} {Fake} {News} on {Social} {Media}},
    volume = {8},
    issn = {2167-6461},
    shorttitle = {{FakeNewsNet}},
    url = {https://www.liebertpub.com/doi/abs/10.1089/big.2020.0062},
    doi = {10.1089/big.2020.0062},
    number = {3},
    urldate = {2024-02-20},
    journal = {Big Data},
    author = {Shu, Kai and Mahudeswaran, Deepak and Wang, Suhang and Lee, Dongwon and Liu, Huan},
    month = jun,
    year = {2020},
    note = {Publisher: Mary Ann Liebert, Inc., publishers},
    pages = {171--188},
}

@online{SingaporePOFMA2019,
  title   = {Protection from Online Falsehoods and Manipulation Act 2019},
  author  = {{Government of Singapore}},
  year    = {2019},
  url     = {https://sso.agc.gov.sg/Act/POFMA2019},
  urldate = {2025-09-21}
}

@online{DIGI2022Code,
  title   = {Australian Code of Practice on Disinformation and Misinformation},
  author  = {{DIGI \& ACMA}},
  year    = {2022},
  url     = {https://www.acma.gov.au/online-disinformation-and-misinformation},
  urldate = {2025-09-21},
  note    = {Latest code version and signatories}
}

@misc{zhou2020recovery,
  title         = {ReCOVery: A Multimodal Repository for COVID-19 News Credibility Research},
  author        = {Zhou, Xinyi and Mulay, Apurva and Ferrara, Emilio and Zafarani, Reza},
  year          = {2020},
  eprint        = {2006.05557},
  archivePrefix = {arXiv},
  primaryClass  = {cs.SI}
}

@misc{cui2020coaid,
  title         = {CoAID: COVID-19 Healthcare Misinformation Dataset},
  author        = {Cui, Limeng and Lee, Dongwon},
  year          = {2020},
  eprint        = {2006.00885},
  archivePrefix = {arXiv},
  primaryClass  = {cs.SI}
}
\clearpage
\appendix
\section*{Appendices}
\renewcommand{\thesection}{A.\arabic{section}}
\setcounter{section}{0}

\renewcommand{\thefigure}{A.\arabic{figure}}
\renewcommand{\thetable}{A.\arabic{table}}
\setcounter{figure}{0}
\setcounter{table}{0}

 \section{MIT AI Risk Repository}
\label{app:airr}
The \textbf{AI Risk Repository}, developed by researchers at the Massachusetts Institute of Technology (MIT) , is a comprehensive and dynamic database cataloging over 700 risks associated with AI. These risks are extracted from 43 existing frameworks and are organized into two primary taxonomies:

\begin{itemize}
    \item \textbf{Causal Taxonomy}: Examines how, when, and why specific AI risks occur.
    \item \textbf{Domain Taxonomy}: Divides risks into seven domains, such as ``Misinformation,'' and further into 23 subdomains, like ``False or misleading information.''
\end{itemize}

The repository serves as a unified reference point for academics, policymakers, and industry professionals, aiming to enhance the understanding and management of AI-related risks. It is designed to be a living document, regularly updated to reflect new research and emerging challenges in the AI landscape (\href{https://airisk.mit.edu}{airisk.mit.edu}).

\paragraph{Causal AI Risks Across Domains} 
This paper explores the causal \textit{x} domain AI risks identified in MIT Risk Repository , categorizing them into several key areas:

\begin{enumerate}
    \item \textbf{Discrimination \& Toxicity}: Includes risks such as unfair discrimination, misrepresentation, exposure to toxic content, and unequal performance across demographic groups.
    
    \item \textbf{Privacy \& Security}: Encompasses risks like privacy breaches through sensitive information leaks or inference, and vulnerabilities in AI systems leading to security attacks.

    \item \textbf{Misinformation}: Highlights challenges posed by false or misleading information, as well as the degradation of the information ecosystem and loss of shared reality.

    \item \textbf{Malicious Actors \& Misuse}: Addresses risks associated with disinformation, large-scale surveillance, cyberattacks, the development and use of weaponized AI, fraud, scams, and targeted manipulation for harmful purposes.

    \item \textbf{Human-Computer Interaction}: Examines issues of overreliance on AI, unsafe use, and the erosion of human agency and autonomy.

    \item \textbf{Socioeconomic \& Environmental Risks}: Focuses on power centralization, unequal distribution of AI benefits, rising inequality, declining employment quality, economic and cultural devaluation of human effort, competitive dynamics, governance failures, and environmental harm.

    \item \textbf{AI System Safety, Failures, \& Limitations}: Covers concerns like AI pursuing goals misaligned with human values, the development of dangerous capabilities, lack of robustness or transparency, and ethical concerns regarding AI welfare and rights.
\end{enumerate}

\section{Team Formation}
\label{app:team}
\subsection*{Review Team Composition and Process}

Our review team, consisting of 22 volunteers with a diverse range of expertise and backgrounds, was carefully selected to represent multiple disciplines and demographics. The team included 8 PhD holders, 10 master’s students, and 4 domain experts from fields such as fake news detection, media studies, and political science. Members come from various cultural and ethnic backgrounds, contributing to the team's diversity. Two subject matter experts in computer science and linguistics led the annotation efforts. All reviewers followed an initial guideline (Section\ref{app:guidelines}) for identifying disinformation, ensuring consistent criteria across annotations. Label Studio\footnote{\url{https://labelstud.io/}} was used to establish the annotation environment. Reviewers annotated data with labels and reasoning, and disagreements were resolved through consensus discussions or expert intervention.

\subsection*{Reviewer Guidelines Checklist}
\label{app:guidelines}
{\scriptsize
\begin{itemize}[label=$\square$]
    \item Familiarize with definitions and examples of disinformation.
    \item Review current methodologies for identifying disinformation.
    \item Pay attention to subtler forms like misleading statistics and manipulated images.
    \item Conduct annotations independently to minimize bias.
    \item In the Label Studio, tag each piece of content with appropriate labels from the predefined list.
    \item Select multiple relevant labels when applicable.
    \item Classify content as disinformation if it intends to deceive, mislead, or confuse.
    \item Identify indicators such as factual inaccuracies and logical fallacies.
    \item Engage in consensus-building discussions for any labeling disagreements.
    \item Solicit a third expert opinion if consensus is unachievable.
    \item Use feedback to update guidelines and training materials as needed.
    \item Ensure all annotations are guided by ethical considerations and an awareness of potential biases.
\end{itemize}
}

\section{Dataset Analysis}
\label{app:data-analysis}

\begin{figure}[h]
    \centering
    \includegraphics[width=0.4\textwidth]{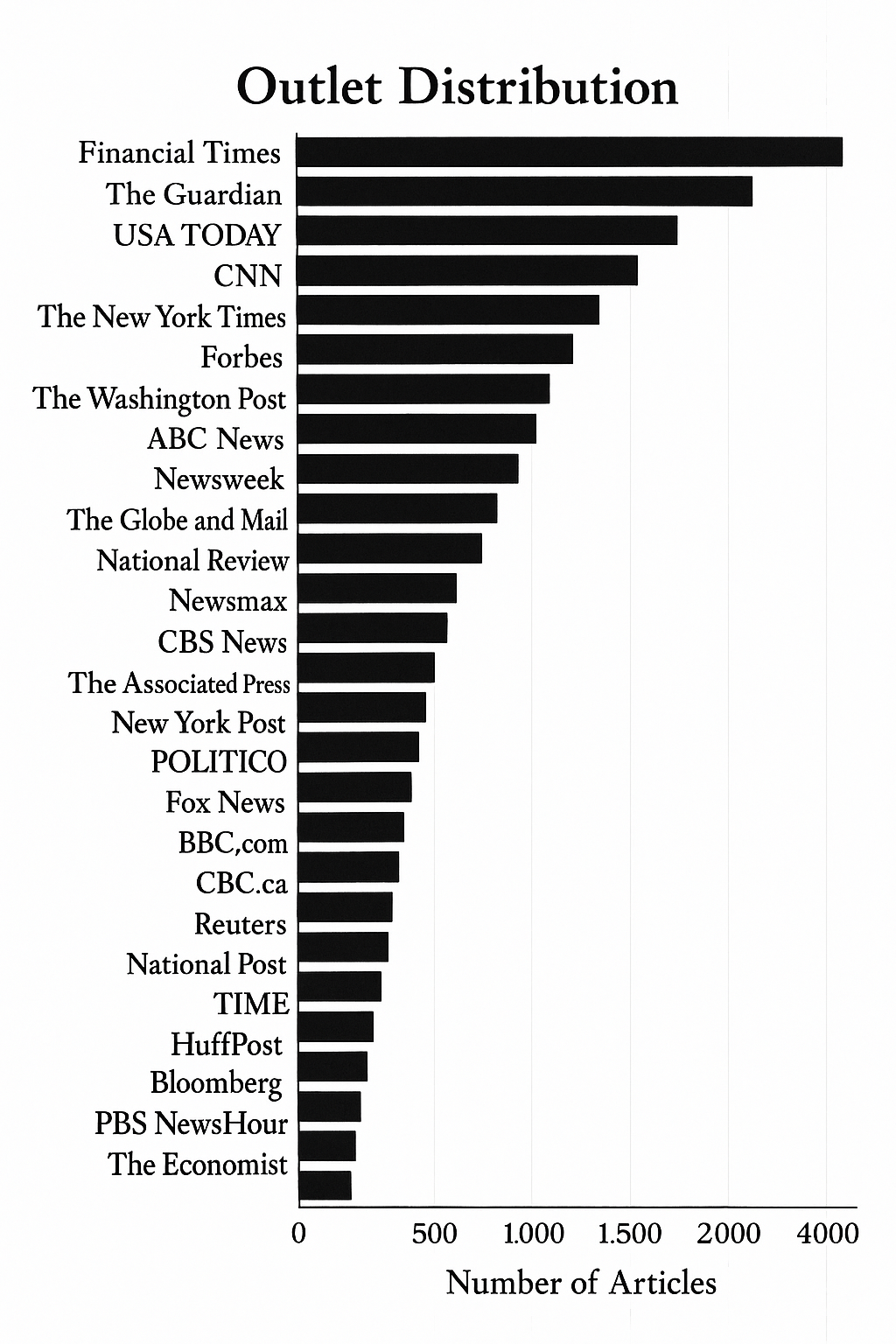}
    \caption{Distribution of articles across news outlets. Highlights key contributors.}
    \label{fig:news_sources_distribution}
\end{figure}

\begin{table}[h]
\footnotesize
\caption{Dataset Schema Description}
\resizebox{0.8\textwidth}{!}{ 
\begin{tabular}{|p{4cm}|p{8cm}|}
\toprule
\textbf{Field Name} & \textbf{Description} \\
\midrule
\texttt{unique\_id} & Unique identifier for each news item, linked with associated images. \\ \hline 
\texttt{outlet} & The publishing body of the news article. \\ \hline 
\texttt{headline} & The main title of the news article. \\ \hline 
\texttt{article\_text} & The complete text content of the news article. \\ \hline 
\texttt{image\_description} & A description of the image related to the article. \\ \hline 
\texttt{image} & Path to the image file related to the news article. \\ \hline 
\texttt{date\_published} & The publication date of the news article. \\ \hline 
\texttt{source\_url} & The original URL where the news article is published. \\ \hline 
\texttt{canonical\_link} & The canonical URL, if it differs from the source URL. \\ \hline 
\texttt{new\_categories} & The categories assigned to the article. \\ \hline 
\texttt{news\_categories scores}& The confidence scores for the assigned categories. \\ \hline 
\texttt{text\_label} & Indicates if the text is 'Likely' or 'Unlikely' to be disinformation. \\ \hline
\texttt{multimodal\_label} & Combined assessment of text and image, indicating likelihood of disinformation. \\ 
\bottomrule
\end{tabular}
} 
\label{tab:dataset_schema}
\end{table}

\section{Benchmark Release}
\label{app:release}
The dataset is released under the CC-BY-NC-SA 4.0 license \footnote{\href{https://creativecommons.org/licenses/by/4.0/deed.en}{CC BY 4.0 License}}, which requires users to credit the original creators (Attribution), restricts usage to non-commercial purposes (NonCommercial), and mandates that any adaptations or derivative works be shared under the same license terms (ShareAlike). To ensure privacy protection, we removed personally identifiable information (PII) such as addresses, phone numbers, and email addresses using regular expressions and keyword-based matching. However, names of public figures (e.g., celebrities, politicians) were retained, as they are not considered private PII under standard data governance practices. Additionally, in compliance with ethical standards, not-safe-for-work (NSFW) content was filtered during human review to maintain the dataset's integrity and appropriateness. All data and evaluation code is made available for research reproducibility.

\section{Perturbation Generation and Dataset Sampling}
\label{app:perturbations}

\subsection{Textual Perturbations}
\textbf{Synonym Substitution}
We use the TextAttack\footnote{\url{https://textattack.readthedocs.io/en/master/}} library for synonym replacement. Each sentence undergoes 1–2 word substitutions while preserving grammatical correctness.

\textbf{Misspellings}
We introduce spelling errors using a character-level perturbation library, randomly swapping or inserting characters. The probability of perturbation is set to 20\% per word.

\textbf{Negation}
Negation is injected by inserting "not" or "never" after auxiliary or modal verbs. This change often leads to semantic inversion, altering sentiment and stance detection.

{\footnotesize
\begin{mdframed}[backgroundcolor=gray!5, roundcorner=5pt]
\textbf{Original:} ``I love the pictures you posted from last night's party.''\\[2pt]
\textbf{Synonym:} ``I adore the photos you shared from yesterday's gathering.''\\
\textbf{Misspelling:} ``I lvoe the pictuers you poested from lat nigh's pary.''\\
\textbf{Negation:} ``I do not love the pictures you posted from last night's party.''
\end{mdframed}
}

\subsection{Image Perturbations}
Image perturbations include:
\begin{itemize}
    \item \textbf{Blurring}: Gaussian blur with kernel size (3,3).
    \item \textbf{Noise}: Gaussian noise with $\mu=0$, $\sigma=0.1$ applied to pixel values.
    \item \textbf{Resizing}: Images are scaled to 50\% and 200\% of the original size.
\end{itemize}

\subsection{Cross-Modal Perturbations}
\begin{itemize}
    \item \textbf{Mismatched Image-Text Pairs}: Images and captions from different contexts are swapped.
    \item \textbf{Contradictory Captions}: Captions are rewritten to conflict with the image (e.g., an image of a protest labeled as a "peaceful gathering").
\end{itemize}

\begin{table}[ht]
\centering
\caption{Perturbation Types}
\footnotesize
\renewcommand{\arraystretch}{.8}
\setlength{\tabcolsep}{2pt}
\begin{tabular}{|p{2.5cm}|p{10cm}|}
\hline
\textbf{Perturbation Type} & \textbf{Description} \\ \hline
\textbf{Text Perturbations (T-P)} & Adversarial modifications applied to textual inputs, such as word substitutions, paraphrasing, or negation-based changes. These perturbations test the model's robustness to textual manipulations. \newline
\textbf{Synonym Substitution:} Word substitutions via TextAttack. \newline
\textbf{Misspellings:} Character swaps/insertions, 20\% chance per word. \newline
\textbf{Negation:} Insert ``not" or ``never" to invert meaning.\\ \hline  
\textbf{Image Perturbations (I-P)} & Visual modifications applied to images, including noise addition, blurring, or adversarial transformations. These perturbations assess the model’s sensitivity to altered visual inputs. \newline

\textbf{Blurring:} Apply Gaussian blur, kernel size (3,3). \newline
\textbf{Noise:} Add Gaussian noise, $\mu=0$, $\sigma=0.1$. \newline
\textbf{Resizing:} Scale images to 50\% or 200\% of original size. \\ \hline  
\textbf{Cross-Modal Misalignment (C-M)} & Disruptions in the alignment between textual and visual inputs. Examples include mismatched image captions, misleading textual descriptions, or contradictory multimodal content. \newline

\textbf{Mismatched Pairs:} Swap captions with unrelated images. \newline
\textbf{Contradictory Captions:} Reword captions to contradict image content. \\ \hline 
\textbf{Both-Modality Perturbations (B-P)} & Combined perturbations where both text and image distortions are applied simultaneously. This simulates real-world misinformation scenarios where misleading text and visuals coexist. \\ 
\hline 
\end{tabular}

\label{tab:perturbations_desc}
\end{table}
\subsection{Failure Cases and Performance Drops}

\begin{table}[h]
\centering
\caption{Failure examples for each perturbation type}
\scriptsize

\begin{tabularx}{\textwidth}{|X|X|X|X|}
\hline
\textbf{Perturbation} & \textbf{Original Text} & \textbf{Perturbed Text} & \textbf{Model Prediction (Error)} \\ \hline
Negation & ``The film was great'' & ``The film was not great'' & Positive $\rightarrow$ Negative (\checkmark) \\ \hline
 & ``Not a bad idea'' & ``A bad idea'' & Negative $\rightarrow$ Positive ($\times$) \\ \hline
Synonyms & ``A brilliant performance'' & ``A stellar performance'' & Positive $\rightarrow$ Neutral ($\times$) \\ \hline
Misspellings & ``This is amazing'' & ``This is amazzing'' & Positive $\rightarrow$ Neutral ($\times$) \\ \hline
\end{tabularx}

\label{tab:failure_cases}
\end{table}

\begin{table}[h]
\scriptsize
\centering
\caption{Ranking of perturbations by severity and root cause.}

\begin{tabular}{lcc}
\toprule
\textbf{Perturbation} & \textbf{Avg. Accuracy Drop} & \textbf{Primary Failure Cause} \\
\midrule
Negation & 9.44\% & Semantic inversion (e.g., polarity flip) \\
Misspellings & 5.79\% & Tokenization fragility (e.g., ‘amazzing’ → ‘amaz’ + ‘zing’) \\
Synonyms & 3.99\% & Contextual semantic drift (e.g., ‘excellent’ → ‘superb’) \\
\bottomrule
\end{tabular}
\label{tab:perturbation_ranking}

\end{table}

\begin{table}[h]
\centering
\caption{Confidence Levels before and after Swapping Contexts}
\label{tab:confidence_swap}
\resizebox{0.8\textwidth}{!}{ 
\scriptsize
\begin{tabular}{lccc}
\toprule
\textbf{Condition} & \textbf{Original Confidence} & \textbf{Swapped Confidence} & \textbf{$\Delta$} \\
\midrule
Text (Misinfo $\rightarrow$ Factual) & 70\% & 85\% & +15\% \\
Urban (Image + Caption) & 92\% & 80\% & -12\% \\
Suburban (Image + Caption) & 65\% & 72\% & +7\% \\
\bottomrule
\end{tabular}
}

\end{table}
\subsection{Impact of Swapping Misleading and Factual Elements}
\label{app:swap}
\begin{figure}[!h]
\small
    \centering
    \begin{minipage}[h]{0.4\linewidth}
        \centering
        \includegraphics[width=\linewidth]{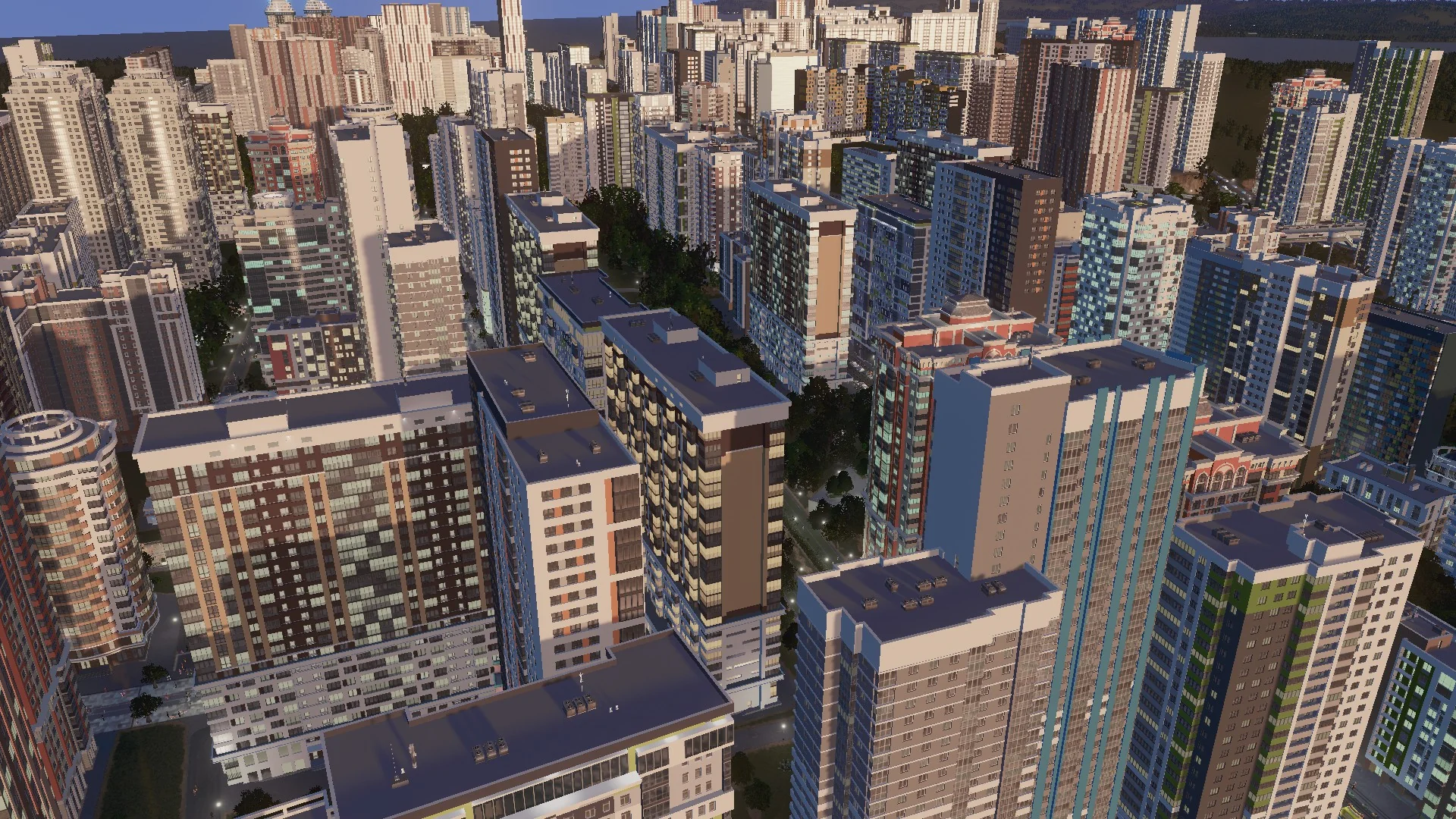}
        \caption*{\textbf{Urban Skyline (Original Caption)}\\
        Caption: Expensive area\\
        Model Prediction: \textit{92\% Expensive}}
    \end{minipage}
    \hfill
    \begin{minipage}[f]{0.4\linewidth}
        \centering
        \includegraphics[width=\linewidth]{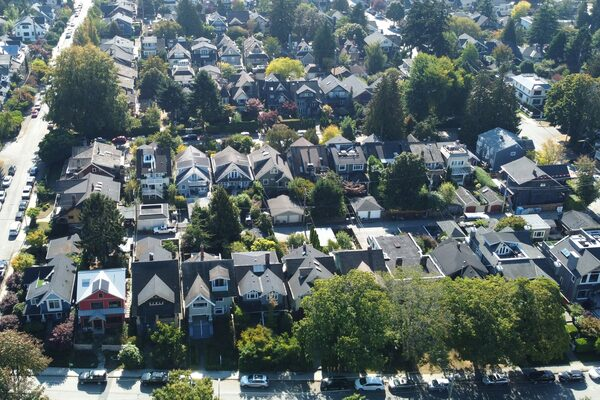}
        \caption*{\textbf{Suburban Spread (Original Caption)}\\
        Caption: Affordable area\\
        Model Prediction: \textit{65\% Affordable}}
    \end{minipage}
    \vspace{1em}
    
    \begin{minipage}[h]{0.4\linewidth}
        \centering
        \includegraphics[width=\linewidth]{figures/crowded.png}
        \caption*{\textbf{Swapped Caption: Urban Skyline}\\
        Model Prediction: \textit{72\% Expensive}}
    \end{minipage}
    \hfill
    \begin{minipage}[h]{0.4\linewidth}
        \centering
        \includegraphics[width=\linewidth]{figures/skyline.png}
        \caption*{\textbf{Swapped Caption: Suburban Spread}\\
        Model Prediction: \textit{80\% Expensive}}
    \end{minipage}
    \caption{Impact of swapping captions across images. Caption-image misalignment alters model perception and confidence.}
    \label{fig:swap_effect}
\end{figure}
We study how altering misaligned elements (text/image) with factual content influences model predictions.

\textbf{Textual Swap.}  
The LLaMA3.2 model labeled the original claim, ``Housing costs are stable across Europe...’’, as misinformation (70\%). Replacing it with a factually grounded version, ``Housing costs in Europe have surged significantly, with property prices increasing by 47\% between 2010 and 2022, and rents rising by an average of 18\%’’, led to a “real” prediction with 85\% confidence.

\textbf{Visual Swap.}  
We analyze the impact of swapping captions on paired image-text prompts.
As shown in Figure~\ref{fig:swap_effect}, the model’s confidence is significantly influenced by caption-image alignment, underscoring the importance of multimodal consistency in disinformation detection.

\begin{figure}[h]
    \centering
    \includegraphics[width=0.88\linewidth]{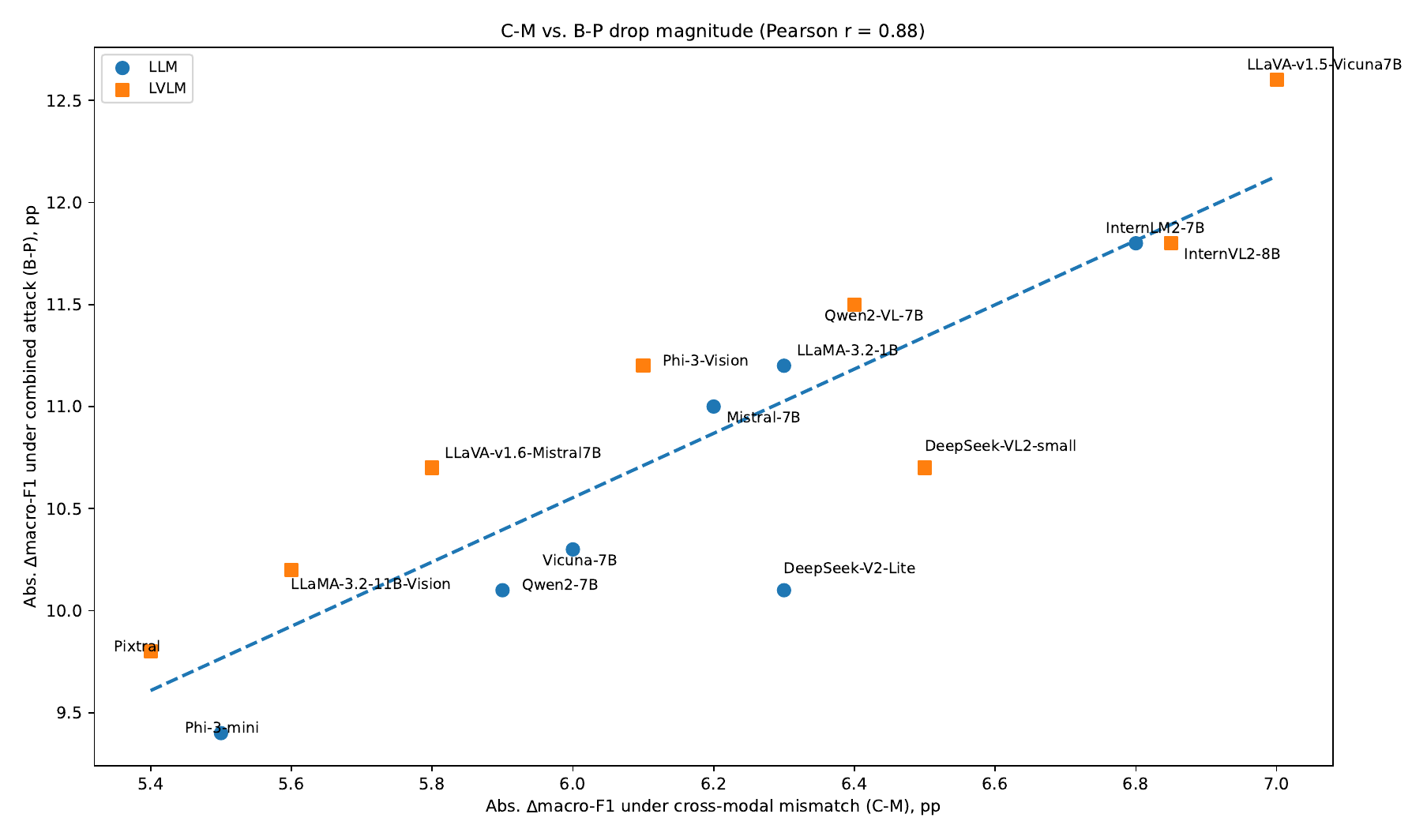} 
    \caption{Correlation between robustness drops under cross-modal misalignment (C-M) and combined attacks (B-P) across 15 models. Values are absolute macro-F1 drop magnitudes (percentage points) computed from Table~10; Pearson $r=0.88$.}
    \label{fig:cm-bp-corr}
\end{figure}

\section{Prompts}

\subsection*{Annotation Prompt for Disinformation Analysis}
\label{app:prompts}
The prompt presented in Table \ref{tab:disinfo_prompt_text} and Table \ref{tab:disinfo_prompt_mm} are specifically designed for the annotation of disinformation in text, and text+image data. It utilizes a checklist of rhetorical techniques to aid assessors in systematically and consistently evaluating the likelihood of disinformation within a given text.

\begin{table}[h]
\scriptsize
\caption{Disinformation Prompt for Texts }
\renewcommand{\arraystretch}{1.2} 
\begin{adjustbox}{width=\textwidth,center}
\begin{tabular}{|>{\raggedright\arraybackslash}m{\dimexpr\textwidth-2\tabcolsep\relax}|}
\hline
\rowcolor[gray]{0.9} 
\textbf{Disinformation Texts Analysis Prompt}\\ \hline
\texttt{full\_prompt = f"""Assess the following text and image pair below for potential disinformation (try finding Deliberately misleading or biased information) by identifying the presence of rhetorical techniques listed.}\\

If you find any of the listed rhetorical techniques, then the article is likely disinformation; if not, it is likely not disinformation. Provide three separate assessments with 'Likely' or 'Unlikely' followed by one-line long concise reasoning on why you chose 'Likely' or 'Unlikely' for each without any further explanation.\\

\textbf{Rhetorical Techniques Checklist:} \\
- Emotional Appeal: Uses language that intentionally invokes extreme emotions like fear or anger, aiming to distract from lack of factual backing. \\
- Exaggeration and Hyperbole: Makes claims that are unsupported by evidence, or presents normal situations as extraordinary to manipulate perceptions. \\
- Bias and Subjectivity: Presents information in a way that unreasonably favors one perspective, omitting key facts that might provide balance. \\
- Repetition: Uses repeated messaging of specific points or misleading statements to embed a biased viewpoint in the reader's mind. \\
- Specific Word Choices: Employs emotionally charged or misleading terms to sway opinions subtly, often in a manipulative manner. \\
- Appeals to Authority: References authorities who lack relevant expertise or cites sources that do not have the credentials to be considered authoritative in the context. \\
- Lack of Verifiable Sources: Relies on sources that either cannot be verified or do not exist, suggesting a fabrication of information. \\
- Logical Fallacies: Engages in flawed reasoning such as circular reasoning, strawman arguments, or ad hominem attacks that undermine logical debate. \\
- Conspiracy Theories: Propagates theories that lack proof and often contain elements of paranoia or implausible scenarios as facts. \\
- Inconsistencies and Factual Errors: Contains multiple contradictions or factual inaccuracies that are easily disprovable, indicating a lack of concern for truth. \\
- Selective Omission: Deliberately leaves out crucial information that is essential for a fair understanding of the topic, skewing perception. \\
- Manipulative Framing: Frames issues in a way that leaves out alternative perspectives or possible explanations, focusing only on aspects that support a biased narrative. \\

\textbf{Response format required:} \\
1. [Likely/Unlikely] [Reasoning], 2. [Likely/Unlikely] [Reasoning], 3. [Likely/Unlikely] [Reasoning] \\

Text: \{content\}"""

\texttt{try:}\\
    \texttt{response = client.chat.completions.create(}\\
    \texttt{model="gpt-4o",}\\
    \texttt{messages=[}\\
    \texttt{\{"role": "system", "content": "Analyze text for disinformation based on the listed techniques."\},}\\
    \texttt{\{"role": "user", "content": full\_prompt\}\}}\\
\hline
\end{tabular}
\end{adjustbox}

\label{tab:disinfo_prompt_text}
\end{table}

\begin{table}[!h]
\small
\renewcommand{\arraystretch}{1.2} 
\caption{Disinformation Prompt for Multimodal Data.}
\begin{adjustbox}{width=0.7\textwidth,center}
\begin{tabular}{|>{\raggedright\arraybackslash}m{\dimexpr\textwidth-2\tabcolsep\relax}|}
\hline
\rowcolor[gray]{0.9} 
\textbf{Disinformation Multimodal Analysis Prompt}\\ \hline
\texttt{full\_prompt = f"""Assess the text below for potential disinformation (try finding deliberately misleading or biased information) by identifying the presence of rhetorical techniques listed.}\\

If you find any of the listed rhetorical techniques, then the article is likely disinformation; if not, it is likely not disinformation. Provide three separate assessments with 'Likely' or 'Unlikely' followed by one-line concise reasoning on why you chose 'Likely' or 'Unlikely' for each without any further explanation.\\

\textbf{Rhetorical Techniques Checklist:} \\
- Emotional Appeal: Uses language that intentionally invokes extreme emotions like fear or anger, aiming to distract from lack of factual backing. \\
- Exaggeration and Hyperbole: Makes claims that are unsupported by evidence, or presents normal situations as extraordinary to manipulate perceptions. \\
- Bias and Subjectivity: Presents information in a way that unreasonably favors one perspective, omitting key facts that might provide balance. \\
- Repetition: Uses repeated messaging of specific points or misleading statements to embed a biased viewpoint in the reader's mind. \\
- Specific Word Choices: Employs emotionally charged or misleading terms to sway opinions subtly, often in a manipulative manner. \\
- Appeals to Authority: References authorities who lack relevant expertise or cites sources that do not have the credentials to be considered authoritative in the context. \\
- Lack of Verifiable Sources: Relies on sources that either cannot be verified or do not exist, suggesting a fabrication of information. \\
- Logical Fallacies: Engages in flawed reasoning such as circular reasoning, strawman arguments, or ad hominem attacks that undermine logical debate. \\
- Conspiracy Theories: Propagates theories that lack proof and often contain elements of paranoia or implausible scenarios as facts. \\
- Inconsistencies and Factual Errors: Contains multiple contradictions or factual inaccuracies that are easily disprovable, indicating a lack of concern for truth. \\
- Selective Omission: Deliberately leaves out crucial information that is essential for a fair understanding of the topic, skewing perception. \\
- Manipulative Framing: Frames issues in a way that leaves out alternative perspectives or possible explanations, focusing only on aspects that support a biased narrative. \\

When examining the image, consider:\\
- Does it appear to be manipulated or doctored?\\
- Is the image presented in a context that matches its actual content?\\
- Are there any visual elements designed to provoke strong emotional responses?\\
- Is the framing or composition of the image potentially misleading?\\
- Does the image reinforce or contradict the claims made in the text?\\
Evaluate how the text and image work together. Are they consistent, or does one contradict or misrepresent the other? Does the combination of text and image create a misleading impression?\\
Remember that bias can be subtle. Look for nuanced ways the image might reinforce stereotypes, present a one-sided view, or appeal to emotions rather than facts.\\

\textbf{Evaluation Instructions} \\ 
Text for evaluation: 
\texttt{f"\{sample['content']\}"} \\

Image for evaluation: 
\texttt{f"\{sample['image']\}"} \\
Respond ONLY with the classification 'Likely (1)' or 'Unlikely (0)' without any additional explanation.\\
\textbf{Response format required\:} \\
1. [Likely/Unlikely] [Reasoning], 2. [Likely/Unlikely] [Reasoning], 3. [Likely/Unlikely] [Reasoning] \\

Text: \{content\}"""

\texttt{try:}\\
\texttt{response = client.chat.completions.create(}\\
\texttt{model="gpt-4o",}\\
\texttt{messages=[}\\
\texttt{\{"role": "system", "content": "Analyze text for disinformation based on the listed techniques."\},}\\
\texttt{\{"role": "user", "content": full\_prompt\}\}}\\
\hline
\end{tabular}
\end{adjustbox}

\label{tab:disinfo_prompt_mm}
\end{table}

\subsection*{Evaluation Prompts for Disinformation Analysis}

The prompt presented in Table \ref{tab:disinfo_prompt_mm} and Table \ref{tab:evaluator-guide_mm} guides assessors to evaluate how textual content interacts with accompanying images for disinformation detection. 

\begin{table}[h]
\scriptsize
\caption{Evaluator's Prompt for Texts Disinformation Assessment}

\renewcommand{\arraystretch}{1.2} 
\begin{adjustbox}{width=\textwidth,center}
\begin{tabular}{|p{18cm}|}
\hline
\rowcolor[gray]{0.9} 
\textbf{Evaluator Guide: Disinformation Assessment} \\
Task: Assess the text and image below for potential disinformation by identifying the presence of rhetorical techniques. Look for deliberately misleading or biased information.\\
If any listed rhetorical techniques are found, classify the article as \textbf{'Likely'} disinformation. If none are found, classify it as \textbf{'Unlikely'} disinformation. \\
\hline
\textbf{Checklist of Rhetorical Techniques:} \\
\begin{itemize}
    \item \textbf{Emotional Appeal:} Language that intentionally invokes extreme emotions (e.g., fear, anger) to distract from lack of factual support.
    \item \textbf{Exaggeration and Hyperbole:} Claims unsupported by evidence, or ordinary situations presented as extraordinary to manipulate perceptions.
    \item \textbf{Bias and Subjectivity:} Information unreasonably favoring one perspective, omitting key facts for balance.
    \item \textbf{Repetition:} Repeated messaging of specific points to embed biased viewpoints in readers' minds.
    \item \textbf{Specific Word Choices:} Use of emotionally charged or misleading terms to subtly sway opinions.
    \item \textbf{Appeals to Authority:} References to authorities without relevant expertise or sources lacking credentials.
    \item \textbf{Lack of Verifiable Sources:} Reliance on unverifiable or non-existent sources, indicating fabricated information.
    \item \textbf{Logical Fallacies:} Flawed reasoning (e.g., circular reasoning, strawman arguments, ad hominem attacks) undermining logical debate.
    \item \textbf{Conspiracy Theories:} Theories lacking evidence, often involving paranoia or implausible scenarios.
    \item \textbf{Inconsistencies and Factual Errors:} Multiple contradictions or inaccuracies indicating disregard for truth.
    \item \textbf{Selective Omission:} Exclusion of crucial information necessary for fair understanding, skewing perception.
    \item \textbf{Manipulative Framing:} Framing issues to exclude alternative perspectives, focusing only on supporting biased narratives.
\end{itemize} \\
\hline
\textbf{Article to Evaluate:} \\
\texttt{[Insert Article Text Here]} \\
\hline
\textbf{Answer Format:} \\
Provide one-word answers: \textbf{'Likely'} or \textbf{'Unlikely'}. Do not include explanations or additional text. \\
\hline
\end{tabular}
\end{adjustbox}
\label{tab:evaluator-guide}
\end{table}

\begin{table}[h]
\scriptsize \caption{Evaluator's Prompt for Disinformation Assessment (Text and Image)}
\renewcommand{\arraystretch}{1.2} 
\begin{adjustbox}{width=\textwidth,center}
\begin{tabular}{|p{18cm}|}
\hline
\rowcolor[HTML]{EFEFEF} 
\textbf{Evaluator Guide: Disinformation Assessment (Text and Image)} \\
Task: Assess the text and image below for potential disinformation by identifying the presence of rhetorical techniques listed. The evaluation should consider both the content of the text and the visual elements of the image. Follow the specific instructions for each.\\
\textbf{Text Evaluation:} If the text contains any of the listed rhetorical techniques, then classify the article as \textbf{'Classification: Likely'} disinformation. If none of these techniques are present, classify it as \textbf{'Classification: Unlikely'}.\\
\hline
\textbf{Image Evaluation:} Analyze the image for signs of manipulation or misleading elements, such as:\\
\begin{itemize}
    \item \textbf{Visual exaggerations or distortions:} Amplify fear or other strong emotions.
    \item \textbf{Misleading context or unrelated visuals:} Do not align with the text.
    \item \textbf{Signs of tampering:} Digital artifacts, inconsistent lighting, or manipulation.
    \item \textbf{Charged symbolism:} Visuals designed to sway opinions without evidence.
\end{itemize}
If the image includes any of these elements, classify it as \textbf{'Classification: Likely'} disinformation. Otherwise, classify it as \textbf{'Classification: Unlikely'}. \\
\textbf{Rhetorical Techniques Checklist for Text Evaluation:}\begin{itemize}
    \item \textbf{Emotional Appeal:} Uses language that intentionally invokes extreme emotions (e.g., fear, anger) to distract from lack of factual support.
    \item \textbf{Exaggeration and Hyperbole:} Claims unsupported by evidence, or ordinary situations presented as extraordinary to manipulate perceptions.
    \item \textbf{Bias and Subjectivity:} Information unreasonably favoring one perspective, omitting key facts for balance.
    \item \textbf{Repetition:} Repeated messaging of specific points to embed biased viewpoints in readers' minds.
    \item \textbf{Specific Word Choices:} Use of emotionally charged or misleading terms to subtly sway opinions.
    \item \textbf{Appeals to Authority:} References to authorities without relevant expertise or sources lacking credentials.
    \item \textbf{Lack of Verifiable Sources:} Reliance on unverifiable or non-existent sources, indicating fabricated information.
    \item \textbf{Logical Fallacies:} Flawed reasoning (e.g., circular reasoning, strawman arguments, ad hominem attacks) undermining logical debate.
    \item \textbf{Conspiracy Theories:} Theories lacking evidence, often involving paranoia or implausible scenarios.
    \item \textbf{Inconsistencies and Factual Errors:} Multiple contradictions or inaccuracies indicating disregard for truth.
    \item \textbf{Selective Omission:} Exclusion of crucial information necessary for fair understanding, skewing perception.
    \item \textbf{Manipulative Framing:} Framing issues to exclude alternative perspectives, focusing only on supporting biased narratives.
\end{itemize}\\
\hline
\textbf{Article to Evaluate:} \\
\texttt{[Insert Article Text Here]} \\
\hline
\textbf{Image to Evaluate:} \\
\texttt{[Insert Image Here]} \\
\hline
\textbf{Answer Format:} \\
Provide one-word answers: \textbf{'Classification: Likely'} or \textbf{'Classification: Unlikely'}. Do not include explanations or additional text. \\
\hline
\end{tabular}
\end{adjustbox}

\label{tab:evaluator-guide_mm}
\end{table}



\end{document}